\documentclass[10pt,twocolumn,letterpaper]{article}
\usepackage{iccv}
\usepackage{times}
\usepackage{epsfig}
\usepackage{graphicx}
\usepackage{amsmath}
\usepackage{amssymb}
\usepackage{color}
\usepackage{multirow}
\usepackage{tabularx}

\usepackage[breaklinks=true,bookmarks=false]{hyperref}

\usepackage{wrapfig}
\usepackage{pdfpages}
\usepackage{sidecap}
\usepackage{colortbl}
\usepackage{overpic}
\usepackage{booktabs}
\usepackage{colortbl}
\usepackage[toc,page]{appendix}
\usepackage{placeins}

\usepackage{rotating}
\usepackage[export]{adjustbox}
\usepackage[table]{xcolor}

\definecolor{red}{rgb}{0.95,0.4,0.4}
\definecolor{blue}{rgb}{0.4,0.4,0.95}
\definecolor{darkblue}{rgb}{0,0,0.8}
\definecolor{darkred}{rgb}{0.8,0,0}
\definecolor{darkgreen}{rgb}{0,0.5,0}
\definecolor{grey}{rgb}{0.6,0.6,0.6}
\definecolor{col1}{RGB}{232, 161, 148}
\definecolor{col2}{RGB}{148, 187, 232}

\usepackage{xspace}
\newcommand*{\ea}{et al.\@\xspace}

\usepackage{pifont}%

\iccvfinalcopy

\def\iccvPaperID{3670} %

\ificcvfinal\fi

\begin{document}

\title{Digging Into Self-Supervised Monocular Depth Estimation}

\author{Cl\'ement Godard$^{1}$\hspace{20pt}Oisin Mac Aodha$^{2}$\hspace{20pt}Michael Firman$^{3}$\hspace{20pt}Gabriel Brostow$^{3,1}$\\$^{1}$UCL \hspace{30pt} $^{2}$Caltech \hspace{30pt} $^{3}$Niantic\\\url{www.github.com/nianticlabs/monodepth2}}

\maketitle

\ificcvfinal\thispagestyle{empty}\fi

\begin{abstract}

Per-pixel ground-truth depth data is challenging to acquire at scale. 
To overcome this limitation, self-supervised learning has emerged as a promising alternative for training models to perform monocular depth estimation.  
In this paper, we propose a set of improvements, which together result in both quantitatively and qualitatively improved depth maps compared to competing self-supervised methods.

Research on self-supervised monocular training usually explores increasingly complex architectures, loss functions, and image formation models, all of which have recently helped to close the gap with fully-supervised methods.
We show that a surprisingly simple model, and associated design choices, lead to superior predictions. 
In particular, we propose 
(i) a minimum reprojection loss, designed to robustly handle occlusions,
(ii) a full-resolution multi-scale sampling method that reduces visual artifacts, and 
(iii) an auto-masking loss to ignore training pixels that violate camera motion assumptions.
We demonstrate the effectiveness of each component in isolation, and show high quality, state-of-the-art results on the KITTI benchmark. 

\end{abstract}

\section{Introduction}
We seek to automatically infer a dense depth image from a single color input image.
Estimating absolute, or even relative depth, seems ill-posed without a second input image to enable triangulation. Yet, humans learn from navigating and interacting in the real-world, enabling us to hypothesize plausible depth estimates for novel scenes \cite{hochberg1952familiar}.

Generating high quality depth-from-color is attractive because it could inexpensively complement LIDAR sensors used in self-driving cars, and enable new single-photo applications such as image-editing and AR-compositing. Solving for depth is also a powerful way to use large unlabeled image datasets for the pretraining of deep networks for downstream discriminative tasks~\cite{jiang2017self}.
However, collecting large and varied training datasets with accurate \emph{ground truth} depth for supervised learning \cite{saxena2009make3d,eigen2014depth} is itself a formidable challenge.
As an alternative, several recent self-supervised approaches have shown that it is instead possible to train monocular depth estimation models using only synchronized \emph{stereo pairs} ~\cite{garg2016unsupervised,godard2017unsupervised} or \emph{monocular video} \cite{zhou2017unsupervised}. %

\begin{figure}[!t]
  \centering

\newcommand{\turnwidth}{0.485\columnwidth}

\newcommand{\imlabel}[2]{\includegraphics[width=0.49\columnwidth]{#1}%
\raisebox{2pt}{\makebox[-2pt][r]{\footnotesize #2}}}

\begin{tabular}{@{\hskip 0mm}c@{\hskip 1.5mm}c}

\centering
\setlength\tabcolsep{0.0pt} %

\imlabel{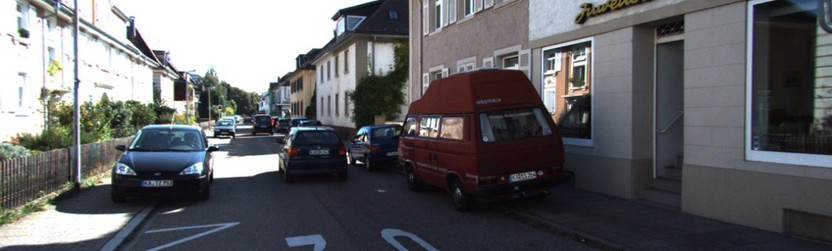}
{\textcolor{white}{Input}} &
\imlabel{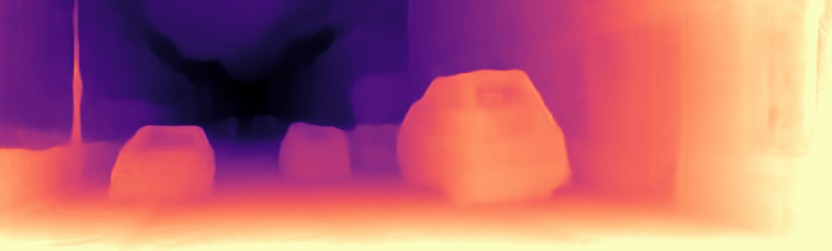}
{\textbf{Monodepth2 (M)}} \\

\imlabel{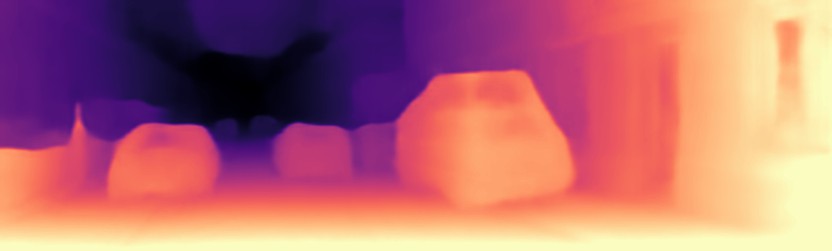}
{\textbf{Monodepth2 (S)}} &
\imlabel{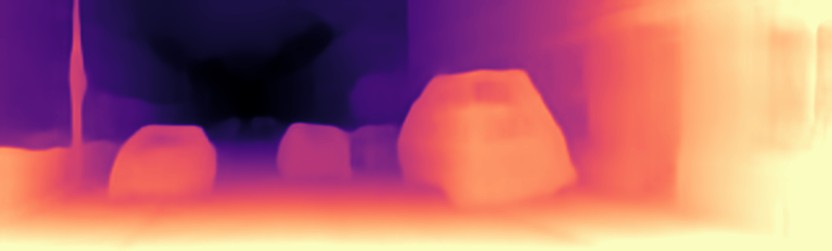}
{\textbf{Monodepth2 (MS)}} \\

\hline
\addlinespace

\imlabel{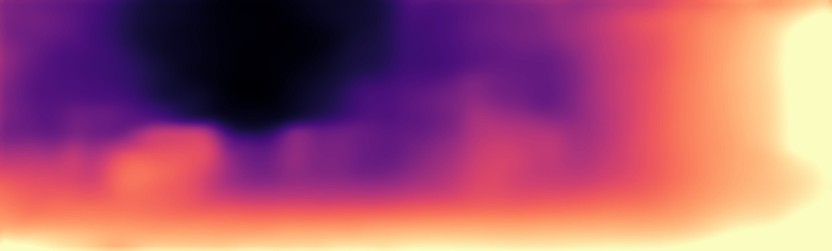}
{Zhou \ea~\cite{zhou2017unsupervised} (\textbf{M})} 
&
\imlabel{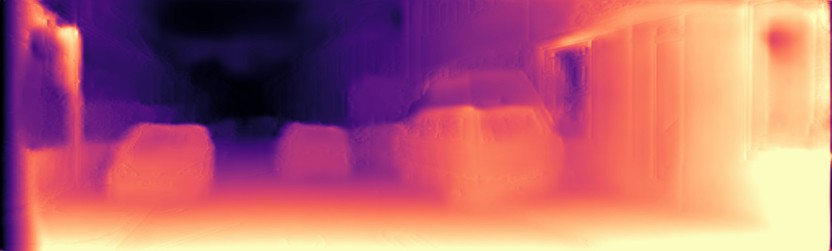}
{Monodepth \cite{godard2017unsupervised} (\textbf{S})} 
\\

\imlabel{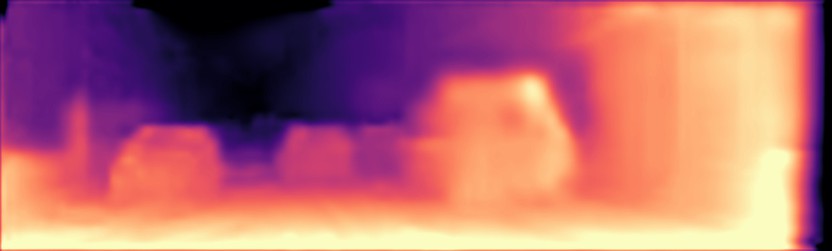}
{Zhan \ea~\cite{zhanst2018} (\textbf{MS})} 
&
\imlabel{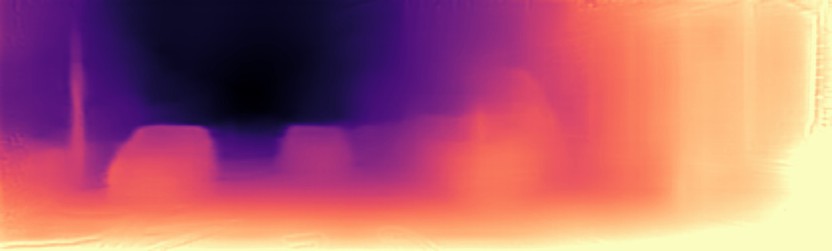}
{DDVO~\cite{wang2017learning} (\textbf{M})} 
\\

\imlabel{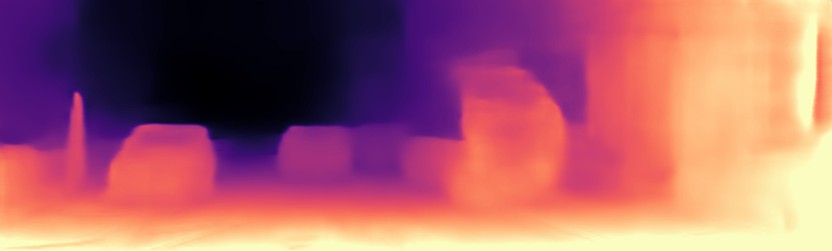}
{Ranjan~\ea~\cite{ranjan2018adversarial} (\textbf{M})} 
&
\imlabel{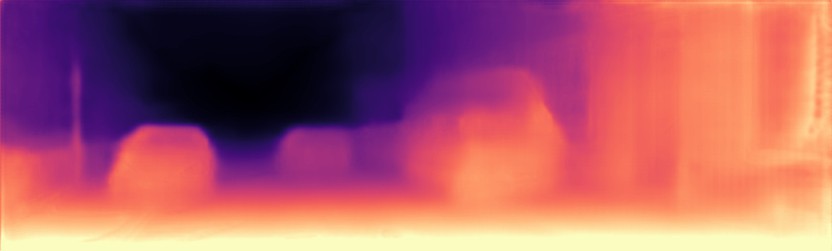}
{EPC++ \cite{luo2018every} (\textbf{MS})} 
\\

\end{tabular}

  \vspace{0pt}
  \caption{{\bf Depth from a single image.} Our self-supervised model, \textbf{Monodepth2}, produces sharp, high quality depth maps, whether trained with monocular (M), stereo (S), or joint (MS) supervision.
  }
  \vspace{-14pt}
  \label{fig:overview}
\end{figure}

Among the two self-supervised approaches, monocular video is an attractive alternative to stereo-based supervision, but it introduces its own set of challenges.
In addition to estimating depth, the model also needs to estimate the egomotion between temporal image pairs during training.
This typically involves training a pose estimation network that takes a finite sequence of frames as input, and outputs the corresponding camera transformations.
Conversely, using stereo data for training makes the camera-pose estimation a one-time offline calibration, but can cause issues related to occlusion and texture-copy artifacts \cite{godard2017unsupervised}.

We propose three architectural and loss innovations that combined, lead to large improvements in monocular depth estimation when training with monocular video, stereo pairs, or both:
(1) A novel appearance matching loss to address the problem of occluded pixels that occur when using monocular supervision.
(2) A novel and simple \emph{auto-masking} approach to ignore pixels where no relative camera motion is observed in monocular training. 
(3) A multi-scale appearance matching loss that performs all image sampling at the input resolution, leading to a reduction in depth artifacts.
Together, these contributions yield state-of-the-art monocular and stereo self-supervised depth estimation results on the KITTI dataset~\cite{Geiger2012CVPR}, and simplify many components found in the existing top performing models.

\begin{figure}
  \centering
  \resizebox{\columnwidth}{!}{
  \newcommand{\imlabel}[2]{\includegraphics[width=0.49\columnwidth]{#1}%
\raisebox{2pt}{\makebox[-2pt][r]{\footnotesize #2}}}

\centering

\begin{tabular}{@{\hskip 1mm}c@{\hskip 1mm}c}

\imlabel{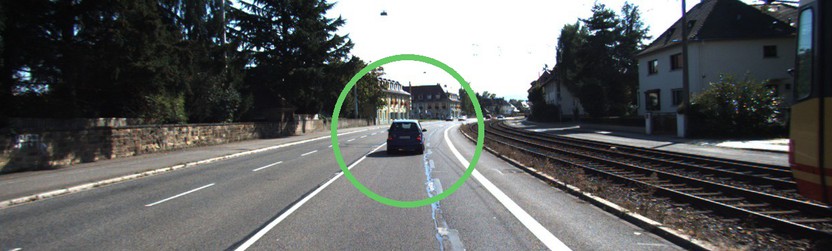}{\textcolor{white}{Input}} &
\imlabel{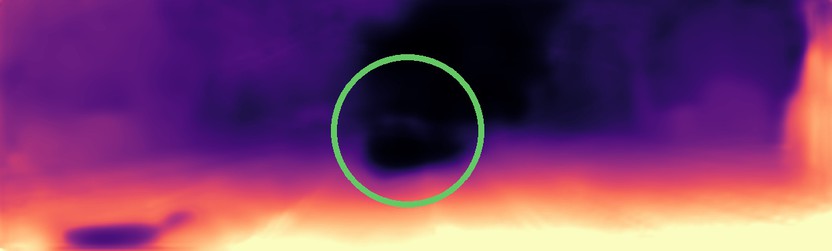}{Geonet \cite{geonet2018} (\textbf{M})} \\
\imlabel{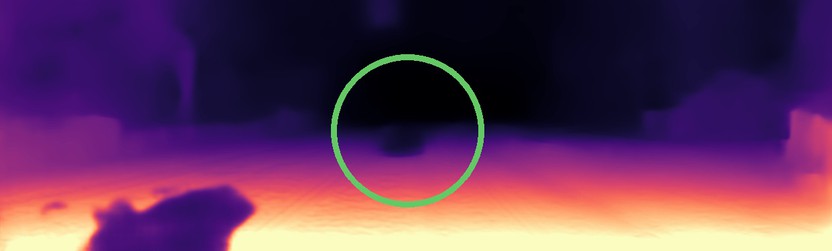}{Ranjan \cite{ranjan2018adversarial} (\textbf{M})} &
\imlabel{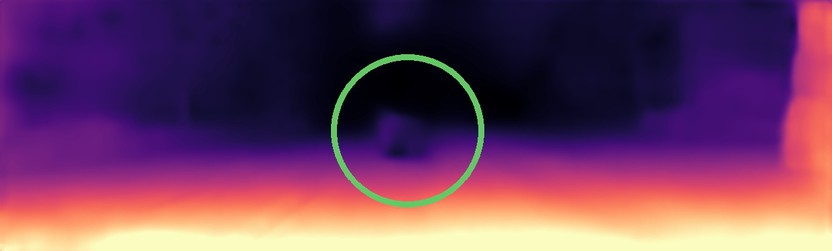}{EPC++ \cite{luo2018every} (\textbf{MS})} \\
\imlabel{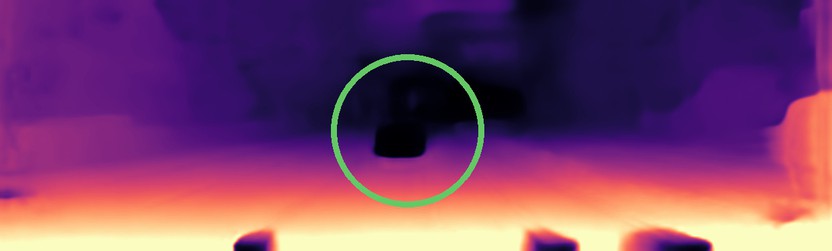}{Baseline (\textbf{M})} &  
\imlabel{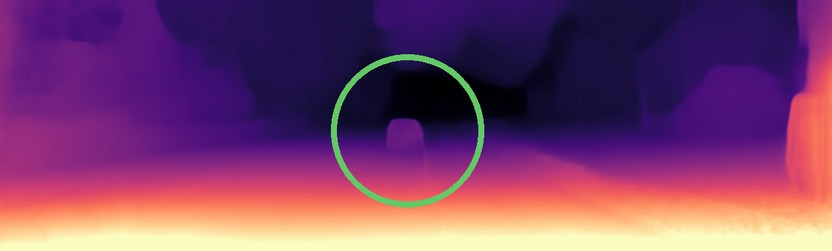}{Monodepth2 (\textbf{M})} \\

\end{tabular}
}
  \caption{{\bf Moving objects.} Monocular methods can fail to predict depth for objects that were often observed to be in motion during training \eg moving cars -- including methods which explicitly model motion \cite{geonet2018,luo2018every,ranjan2018adversarial}. Our method succeeds here where others, and our baseline with our contributions turned off, fail.}
  \label{fig:motion_fail}
  \vspace{-6pt}
\end{figure}

\section{Related Work}
\vspace{-5pt}
We review models that, at test time, take a single color image as input and predict the depth of each pixel as output.

\vspace{-3pt}
\subsection{Supervised Depth Estimation}
\vspace{-5pt}
Estimating depth from a single image is an inherently ill-posed problem as the same input image can project to multiple plausible depths.
To address this, learning based methods have shown themselves capable of fitting predictive models that exploit the relationship between color images and their corresponding depth.
Various approaches, such as combining local predictions \cite{hoiem2005automatic,saxena2009make3d}, non-parametric scene sampling \cite{karsch2014depth}, through to end-to-end supervised learning \cite{eigen2014depth,laina2016deeper,fu2018deep} have been explored.
Learning based algorithms are also among some of the best performing for stereo estimation~ \cite{vzbontar2016stereo,mayer2015large,ummenhofer2017demon,kendall2017end} and optical flow~ \cite{ilg2017flownet,wang2017occlusion}.

Many of the above methods are fully supervised, requiring ground truth depth during training.
However, this is challenging to acquire in varied real-world settings.
As a result, there is a growing body of work that exploits weakly supervised training data, \eg in the form of known object sizes \cite{wu2018size}, sparse ordinal depths \cite{zoran2015learning,chen2016single}, supervised appearance matching terms \cite{vzbontar2016stereo,zhanst2018}, or unpaired synthetic depth data \cite{gandepth2018,atapour2018real,guo2018learning,zou2018df}, all while still requiring the collection of additional depth or other annotations.
Synthetic training data is an alternative \cite{mayer2018makes}, but it is not trivial to generate large amounts of synthetic data containing varied real-world appearance and motion.
Recent work has shown that conventional structure-from-motion (SfM) pipelines can generate sparse training signal for both camera pose and depth  \cite{li2018megadepth,klodt2018supervising,yang2018deep}, where SfM is typically run as a pre-processing step decoupled from learning.
Recently, \cite{depthHints2019} built upon our model by incorporating noisy depth hints from traditional stereo algorithms, improving depth predictions.   

\vspace{-3pt}
\subsection{Self-supervised Depth Estimation}
\vspace{-5pt}
In the absence of ground truth depth, one alternative is to train depth estimation models using image reconstruction as the supervisory signal.
Here, the model is given a set of images as input, either in the form of stereo pairs or monocular sequences. By hallucinating the depth for a given image and projecting it into nearby views, the model is trained by minimizing the image reconstruction error.

\begin{figure*}[t]
  \centering
    \includegraphics[width=1.0\textwidth, page=1]{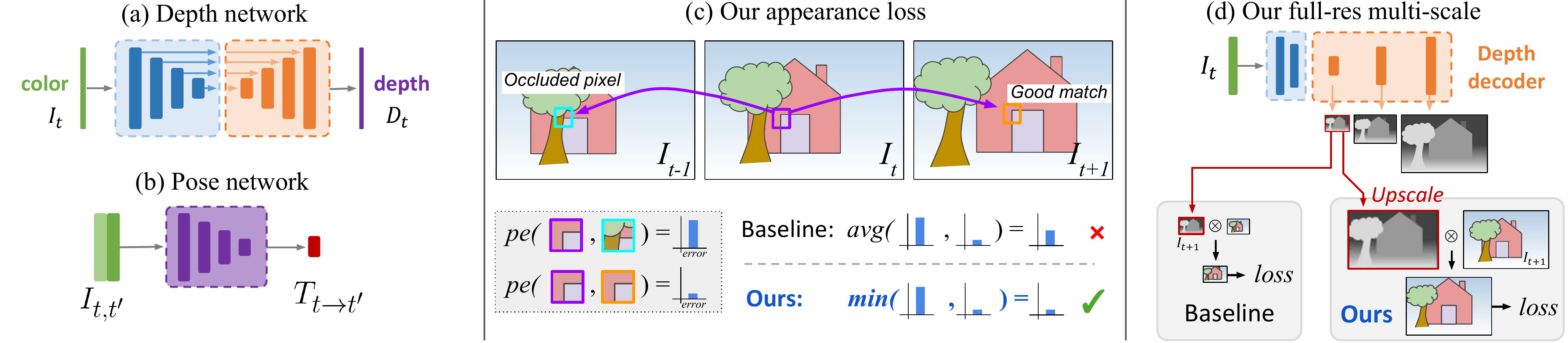}
    \vspace{-12pt}
    \caption{\textbf{Overview.}
    \textbf{(a) Depth network:} We use a standard, fully convolutional, U-Net to predict depth.
    \textbf{(b) Pose network:} Pose between a pair of frames is predicted with a separate pose network.
    \textbf{(c) Per-pixel minimum reprojection:} When correspondences are \emph{good}, the reprojection loss should be \emph{low}.
    However, occlusions and disocclusions result in pixels from the current time step not appearing in both the previous and next frames.
    The baseline \emph{average} loss forces the network to match occluded pixels, whereas our \emph{minimum reprojection} loss only matches each pixel to the view in which it is visible, leading to sharper results.
    \textbf{(d) Full-resolution multi-scale:} We upsample depth predictions at intermediate layers and compute all losses at the input resolution, reducing texture-copy artifacts.
    }%
    \vspace{-4pt}
    \label{fig:model_arch}
\end{figure*}

\vspace{5pt}
\noindent{\textbf{Self-supervised Stereo Training}}\\
\indent{}
One form of self-supervision comes from stereo pairs.
Here, synchronized stereo pairs are available during training, and by predicting the pixel disparities between the pair, a deep network can be trained to perform monocular depth estimation at test time.
\cite{xie2016deep3d} proposed such a model with discretized depth for the problem of novel view synthesis.
\cite{garg2016unsupervised} extended this approach by predicting continuous disparity values, and \cite{godard2017unsupervised} produced results superior to contemporary supervised methods by including a left-right depth consistency term.
Stereo-based approaches have been extended with semi-supervised data \cite{kuznietsov2017semi,singlestereo2018}, generative adversarial networks \cite{aleotti2018generative, pilzer2018unsupervised}, additional consistency \cite{poggi20183net}, temporal information \cite{li2017undeepvo,zhanst2018,babu2018undemon}, and for real-time use~\cite{poggi2018towards}.

In this work, we show that with careful choices regarding appearance losses and image resolution, we can reach the performance of stereo training using only monocular training.
Further, one of our contributions carries over to stereo training, resulting in increased performance there too.

\vspace{5pt}
\noindent{\textbf{Self-supervised Monocular Training}}\\
\indent{}
A less constrained form of self-supervision is to use monocular videos, where consecutive temporal frames provide the training signal.
Here, in addition to predicting depth, the network has to also estimate the camera pose between frames, which is challenging in the presence of object motion.
This estimated camera pose is only needed during training to help constrain the depth estimation network.

In one of the first monocular self-supervised approaches, \cite{zhou2017unsupervised} trained a depth estimation network along with a separate pose network.
To deal with non-rigid scene motion, an additional motion explanation mask allowed the model to ignore specific regions that violated the rigid scene assumption.
However, later iterations of their model available online disabled this term, achieving superior performance.
Inspired by \cite{byravan2017se3}, \cite{vijayanarasimhan2017sfm} proposed a more sophisticated motion model using multiple motion masks.
However, this was not fully evaluated, making it difficult to understand its utility. 
\cite{geonet2018} also decomposed motion into rigid and non-rigid components, using depth and optical flow to explain object motion.
This improved the \emph{flow} estimation, but they reported no improvement when jointly training for flow and depth estimation.
In the context of optical flow estimation, \cite{janai2018unsupervised} showed that it helps to explicitly model occlusion.

Recent approaches have begun to close the performance gap between monocular and stereo-based self-supervision.
\cite{yang2017unsupervised} constrained the predicted depth to be consistent with predicted surface normals, and \cite{yang2018lego} enforced edge consistency.
\cite{mahjourian2018unsupervised} proposed an approximate geometry based matching loss to encourage temporal depth consistency.
\cite{wang2017learning} use a depth normalization layer to overcome the preference for smaller depth values that arises from the commonly used depth smoothness term from \cite{godard2017unsupervised}.
\cite{casser2018depth} make use of pre-computed instance segmentation masks for known categories to help deal with moving objects.

\vspace{5pt}
\noindent{\textbf{Appearance Based Losses}}\\
\indent{}Self-supervised training typically relies on making assumptions about the appearance (\ie brightness constancy) and material properties (\eg Lambertian) of object surfaces between frames.
\cite{godard2017unsupervised} showed that the inclusion of a local structure based appearance loss \cite{wang2004image} significantly improved depth estimation performance compared to simple pairwise pixel differences \cite{xie2016deep3d,garg2016unsupervised,zhou2017unsupervised}.
\cite{klodt2018supervising} extended this approach to include an error fitting term, and \cite{mehta2018structured} explored combining it with an adversarial based loss to encourage realistic looking synthesized images.
Finally, inspired by \cite{vzbontar2016stereo}, \cite{zhanst2018} use ground truth depth to train an appearance matching term.

\section{Method}
Here, we describe our depth prediction network that takes a single color input $I_t$ and produces a depth map $D_t$. We first review the key ideas behind self-supervised training for monocular depth estimation, and then describe our depth estimation network and joint training loss.

\subsection{Self-Supervised Training}
Self-supervised depth estimation frames the learning problem as one of novel view-synthesis, by training a network to predict the appearance of a target image from the viewpoint of \emph{another} image.
By constraining the network to perform image synthesis using an intermediary variable, in our case depth or disparity, we can then extract this interpretable depth from the model.
This is an ill-posed problem as there is an extremely large number of possible incorrect depths per pixel which can  correctly reconstruct the novel view given the relative pose between those two views.
Classical binocular and multi-view stereo methods typically address this ambiguity by enforcing smoothness in the depth maps, and by computing photo-consistency on patches when solving for per-pixel depth via global optimization \eg~\cite{furukawa2015multi}.

Similar to \cite{garg2016unsupervised,godard2017unsupervised,zhou2017unsupervised}, we also formulate our problem as the minimization of a photometric reprojection error at training time.
We express the relative pose for each source view $I_{t^\prime}$, with respect to the target image $I_t$'s pose, as $T_{t \to t^\prime}$.
We predict a dense depth map $D_t$ that minimizes the photometric reprojection error $L_p$, where
\vspace{-2pt}
\begin{eqnarray}
 \quad L_p &=& \sum_{t^\prime} pe(I_t, I_{t^\prime \to t}),\\
\text{and} \quad I_{t^\prime \to t} &=& I_{t^\prime}\Big\langle proj(D_t, T_{t \to t^\prime}, K) \Big\rangle.
\end{eqnarray}
Here $pe$ is a photometric reconstruction error, \eg the L1 distance in pixel space; $proj()$ are the resulting 2D coordinates of the projected depths $D_t$ in $I_{t^\prime}$ and $\big\langle\big\rangle$ is the sampling operator.
For simplicity of notation we assume the pre-computed intrinsics $K$ of all the views are identical, though they can be different.
Following~\cite{jaderberg2015spatial} we use bilinear sampling to sample the source images, which is locally sub-differentiable, and we follow \cite{lossfunctions,godard2017unsupervised} in using L1 and SSIM \cite{wang2004image} to make our photometric error function $pe$,  \ie
\begin{align*}
\quad pe(I_a, I_b) &= \frac{\alpha}{2} (1 - \mathrm{SSIM}(I_a, I_b)) + (1 - \alpha) \|I_a - I_b\|_1,
\end{align*}
where $\alpha = 0.85$.
As in \cite{godard2017unsupervised} we use edge-aware smoothness
\begin{eqnarray}
L_s &=& \left | \partial_x d^*_t   \right | e^{-\left | \partial_x I_t \right |} + \left | \partial_y d^*_t   \right | e^{-\left | \partial_y I_t \right |},
\end{eqnarray} where $d^*_t = d_t / \overline{d_t}$ is the mean-normalized inverse depth from \cite{wang2017learning} to discourage shrinking of the estimated depth.

In stereo training, our source image $I_{t^\prime}$ is the second view in the stereo pair to $I_t$, which has known relative pose.
While relative poses are not known in advance for monocular sequences, \cite{zhou2017unsupervised} showed that it is possible to train a second pose estimation network to predict the relative poses $T_{t \to t^\prime}$ used in the projection function $proj$.
During training, we solve for camera pose and depth simultaneously, to minimize $L_p$.
For monocular training, we use the two frames temporally adjacent to $I_t$ as our source frames, \ie $I_{t^\prime} \in \{ I_{t-1}, I_{t+1} \}$. 
In mixed training (MS), $I_{t^\prime}$ includes the temporally adjacent frames and the opposite stereo view.

\subsection{Improved Self-Supervised Depth Estimation}
Existing monocular methods produce lower quality depths than the best fully-supervised models.
To close this gap, we propose several improvements that significantly increase predicted depth quality, without adding additional model components that also require training (see Fig.~\ref{fig:model_arch}).

\begin{figure}
  \centering
 \includegraphics[width=0.8\columnwidth]{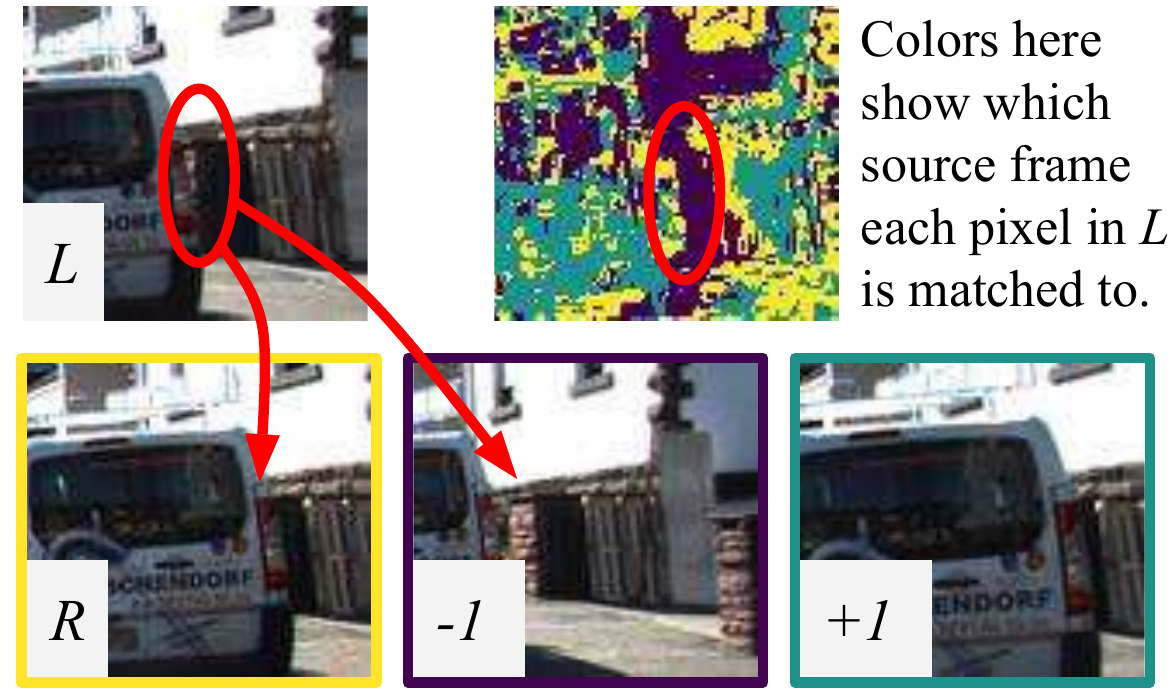}
 \caption{\textbf{Benefit of min.~reprojection loss in MS training}. 
    Pixels in the the circled region are occluded in $I_R$ so no loss is applied between ($I_L$, $I_R$). 
 Instead, the pixels are matched to $I_{-1}$ where they are visible. 
 Colors in the top right image indicate which of the source images on the bottom are selected for matching by Eqn.~\ref{eqn:min_porj}.
 \label{fig:min_reprojection}}
 \vspace{-3pt}
\end{figure}

\vspace{5pt}
\noindent{}{\bf Per-Pixel Minimum Reprojection Loss}\\
When computing the reprojection error from multiple source images, existing self-supervised depth estimation methods average together the reprojection error into each of the available source images.%
This can cause issues with pixels that are visible in the target image, but are \textit{not visible} in some of the source images (Fig.~\ref{fig:model_arch}(c)).
If the network predicts the correct depth for such a pixel, the corresponding color in an occluded source image will likely \emph{not} match the target, inducing a high photometric error penalty.
Such problematic pixels come from two main categories: out-of-view pixels due to egomotion at image boundaries, and occluded pixels.
The effect of out-of-view pixels can be reduced by masking such pixels in the reprojection loss \cite{mahjourian2018unsupervised,vijayanarasimhan2017sfm}, but this does not handle disocclusion, where average reprojection can result in blurred depth discontinuities.

\begin{figure}[!t]
  \centering
  \resizebox{0.95\columnwidth}{!}{
  \newcommand{\turnwidth}{0.49\columnwidth}
\centering

\renewcommand{\arraystretch}{0}
\setlength\tabcolsep{0pt}

\newcolumntype{C}[1]{>{\centering\let\newline\\\arraybackslash\hspace{0pt}}m{#1}}

\begin{tabular}{C{0.5\columnwidth}C{0.5\columnwidth}}

\includegraphics[width=\turnwidth]{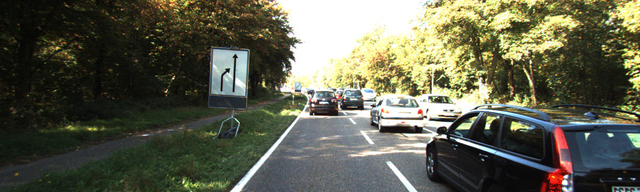} &
\includegraphics[width=\turnwidth]{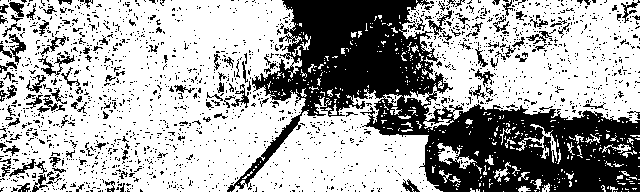} \\
\vspace{2.5pt} \\  
\includegraphics[width=\turnwidth]{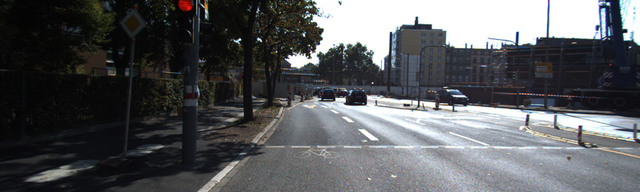} &
\includegraphics[width=\turnwidth]{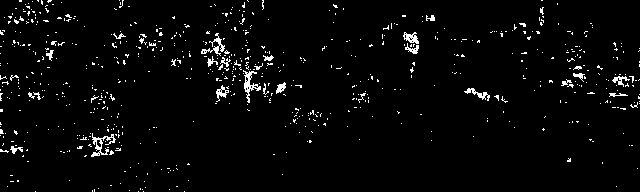} \\

\end{tabular}
}
  \vspace{3pt}
  \caption{{\bf Auto-masking.} We show auto-masks computed after one epoch, where black pixels are removed from the loss (\ie $\mu=0$). 
  The mask prevents objects moving at similar speeds to the camera (top) and whole frames where the camera is static (bottom) from contaminating the loss. The mask is computed from the input frames and network predictions using Eqn.~\ref{eqn:automask}.}
  \label{fig:automasking}
  \vspace{-3pt}
\end{figure}

We propose an improvement that deals with both issues at once.
At each pixel, instead of averaging the photometric error over all source images, we simply use the minimum.
Our final \emph{per-pixel} photometric loss is therefore
\begin{align}
L_p &= \min_{t^\prime} pe(I_t, I_{t^\prime \to t})\label{eqn:min_porj}. %
\end{align}
See Fig.~\ref{fig:min_reprojection} for an example of this loss in practice.
Using our minimum reprojection loss significantly reduces artifacts at image borders, improves the sharpness of occlusion boundaries, and leads to better accuracy (see Table~\ref{tab:kitti_eigen_ablation}).

\vspace{5pt}
\noindent\textbf{Auto-Masking Stationary Pixels} \newline
\noindent Self-supervised monocular training often operates under the assumptions of a moving camera and a static scene. 
When these assumptions break down, for example when the camera is stationary or there is object motion in the scene, performance can suffer greatly.
This problem can manifest itself as `holes' of infinite depth in the predicted test time depth maps, for objects that are typically observed to be moving during training \cite{luo2018every} (Fig.~\ref{fig:motion_fail}).
This motivates our second contribution: a simple auto-masking method that filters out pixels which do not change appearance from one frame to the next in the sequence. This has the effect of letting the network ignore objects which move at the same velocity as the camera, and even to ignore whole frames in monocular videos when the camera stops moving.

Like other works \cite{zhou2017unsupervised, vijayanarasimhan2017sfm, luo2018every}, we also apply a per-pixel mask $\mu$ to the loss, selectively weighting pixels.
However in contrast to prior work, our mask is binary, so $\mu \in \{0, 1\}$, and is computed automatically on the forward pass of the network, instead of being learned or estimated from object motion.
We observe that pixels which remain the same between adjacent frames in the sequence often indicate a static camera, an object moving at equivalent relative translation to the camera, or a low texture region.
We therefore set $\mu$ to only include the loss of pixels where the reprojection error of the warped image $I_{t^\prime \to t}$ is lower than that of the original, unwarped source image $I_t^\prime$, \ie
\begin{align}
    \mu &= \big[ \, 
        \min_{t^\prime} pe(I_t, I_{t^\prime \to t}) 
            < 
        \min_{t^\prime} pe(I_t, I_{t^\prime})     
            \, \big],
            \label{eqn:automask}
\end{align}
where $[\,]$ is the Iverson bracket.
In cases where the camera and another object are both moving at a similar velocity, $\mu$ prevents the pixels which remain stationary in the image from contaminating the loss.
Similarly, when the camera is static, the mask can filter out all pixels in the image (Fig.~\ref{fig:automasking}).
We show experimentally that this simple and inexpensive modification to the loss brings significant improvements.

\vspace{8pt}
\noindent{}{\bf Multi-scale Estimation}\\
Due to the gradient locality of the bilinear sampler \cite{jaderberg2015spatial}, and to prevent the training objective getting stuck in local minima, existing models use multi-scale depth prediction and image reconstruction. Here, the total loss is the combination of the individual losses at each scale in the decoder.
\cite{garg2016unsupervised, godard2017unsupervised} compute the photometric error on images at the resolution of each decoder layer.
We observe that this has the tendency to create `holes' in large low-texture regions in the intermediate lower resolution depth maps, as well as texture-copy artifacts (details in the depth map incorrectly transferred from the color image). %
Holes in the depth can occur at low resolution in low-texture regions where the photometric error is ambiguous.
This complicates the task for the depth network, now freed to predict incorrect depths.

Inspired by techniques in stereo reconstruction \cite{scharstein2002taxonomy}, we propose an improvement to this multi-scale formulation, where we decouple the resolutions of the disparity images and the color images used to compute the reprojection error.
Instead of computing the photometric error on the ambiguous low-resolution images, we first upsample the lower resolution depth maps (from the intermediate layers) to the input image resolution, and then reproject, resample, and compute the error $pe$ at this higher input resolution (Fig.~\ref{fig:model_arch}~(d)).
This procedure is similar to matching patches, as low-resolution disparity values will be responsible for warping an entire `patch' of pixels in the high resolution image.
This effectively constrains the depth maps at each scale to work toward the same objective \ie~reconstructing the high resolution input target image as accurately as possible.

\vspace{5pt}
\noindent{}{\bf Final Training Loss}\\
We combine our per-pixel smoothness and masked photometric losses as $L = \mu L_p + \lambda L_s$, and average over each pixel, scale, and batch.

\begin{table*}[t!]

  \centering
  \resizebox{0.7\textwidth}{!}{
  \begin{tabular}{|l|c||c|c|c|c|c|c|c|}
  \hline
  Method & Train & \cellcolor{col1}Abs Rel & \cellcolor{col1}Sq Rel & \cellcolor{col1}RMSE  & \cellcolor{col1}RMSE log & \cellcolor{col2}$\delta < 1.25 $ & \cellcolor{col2}$\delta < 1.25^{2}$ & \cellcolor{col2}$\delta < 1.25^{3}$\\
  \hline
Eigen \cite{eigen2014depth} & D & 0.203 & 1.548 & 6.307 & 0.282 & 0.702 & 0.890 & 0.890\\
Liu \cite{liu2015learning} & D & 0.201 & 1.584 & 6.471 & 0.273 & 0.680 & 0.898 & 0.967\\
Klodt \cite{klodt2018supervising} & D*M & 0.166 & 1.490 & 5.998 & - &  0.778 & 0.919 & 0.966\\
AdaDepth \cite{gandepth2018}  & D* & 0.167 & 1.257 & 5.578 & 0.237 & 0.771 & 0.922 & 0.971\\
Kuznietsov \cite{kuznietsov2017semi} & DS & 0.113 & 0.741 & 4.621 & 0.189 & 0.862 & 0.960 & 0.986\\
DVSO \cite{yang2018deep} & D*S & 0.097 & 0.734 & 4.442 & 0.187 & 0.888 & 0.958 & 0.980\\
SVSM FT \cite{singlestereo2018} & DS & \underline{0.094} & \underline{0.626} & 4.252 & 0.177 & 0.891 & 0.965 & 0.984\\
Guo \cite{guo2018learning} & DS & 0.096 & 0.641  & \underline{4.095}  & \underline{0.168}  & \underline{0.892}  & \underline{0.967}  & \underline{0.986} \\
DORN \cite{fu2018deep} & D & \textbf{0.072}&  \textbf{0.307} & \textbf{2.727} & \textbf{0.120} & \textbf{0.932} & \textbf{0.984} & \textbf{0.994}\\ 

\arrayrulecolor{black}\hline

Zhou \cite{zhou2017unsupervised}\textdagger & M & 0.183 & 1.595 & 6.709 & 0.270 & 0.734 & 0.902 & 0.959\\
Yang \cite{yang2017unsupervised} & M & 0.182 & 1.481  & 6.501  & 0.267  & 0.725  & 0.906  & 0.963\\
Mahjourian \cite{mahjourian2018unsupervised} & M & 0.163 & 1.240 & 6.220 & 0.250 & 0.762 & 0.916 & 0.968\\

GeoNet \cite{geonet2018}\textdagger & M  & 0.149 & 1.060 & 5.567 & 0.226 & 0.796 & 0.935 & 0.975\\
DDVO \cite{wang2017learning} & M  & 0.151 & 1.257 & 5.583 & 0.228 & 0.810 & 0.936 & 0.974\\
DF-Net \cite{zou2018df} & M & 0.150 & 1.124 & 5.507 & 0.223 & 0.806 & 0.933 & 0.973\\
LEGO \cite{yang2018lego} & M & 0.162 & 1.352 & 6.276 & 0.252 & - & - & - \\
Ranjan \cite{ranjan2018adversarial}  & M & 0.148 & 1.149 & 5.464 & 0.226 & 0.815 & 0.935 & 0.973\\
EPC++ \cite{luo2018every} & M & 0.141 & 1.029 & 5.350 & 0.216 & 0.816 & 0.941 & 0.976\\
Struct2depth `(M)' \cite{casser2018depth}  & M & 0.141 & \underline{1.026} & 5.291 &  0.215 & 0.816 & 0.945 & \underline{0.979}\\

 {\bf Monodepth2} w/o pretraining & M &   
        \underline{0.132} &   1.044 &   \underline{5.142} &   \underline{0.210} &   \underline{0.845} & \underline{0.948} &  0.977 \\
        
\textbf{Monodepth2} & M &   
 {\bf 0.115} &   {\bf 0.903} &   {\bf 4.863} &   {\bf 0.193} &   {\bf 0.877} &   {\bf 0.959} &   {\bf 0.981} \\ 
\arrayrulecolor{gray}\hline
 \textbf{Monodepth2} (1024 $\times$ 320)  &  M &  
 \textbf{0.115} &   \textbf{0.882} &   \textbf{4.701} &   \textbf{0.190} &   \textbf{0.879} &   \textbf{0.961} &   \textbf{0.982} \\ 

\arrayrulecolor{black}\hline

Garg \cite{garg2016unsupervised}\textdagger & S  &  0.152 & 1.226 & 5.849 & 0.246 & 0.784 & 0.921 & 0.967\\
Monodepth R50 \cite{godard2017unsupervised}\textdagger & S & 0.133 & 1.142 & 5.533 & 0.230 & 0.830 & 0.936 & 0.970\\
StrAT \cite{mehta2018structured}  & S  &  0.128 & 1.019 & 5.403 & 0.227 & 0.827 & 0.935 & 0.971\\
3Net  (R50) \cite{poggi20183net} & S & 0.129 & 0.996 & 5.281 & 0.223 & 0.831 & 0.939 & 0.974 \\
3Net (VGG)  \cite{poggi20183net} & S & 0.119 &  1.201 & 5.888 & 0.208 &  0.844 & 0.941 & \textbf{0.978}  \\
SuperDepth + pp \cite{pillai2018superdepth} (1024 $\times$ 382)  & S & \underline{0.112} & \underline{0.875} & \textbf{4.958} & \textbf{0.207} & \underline{0.852} & \underline{0.947} & \underline{0.977} \\

{\bf Monodepth2} w/o pretraining & S  &   
0.130 &   1.144 &   5.485 &   0.232 &   0.831 &   0.932 &   0.968 \\ 
{\bf Monodepth2}      &         S & 
\textbf{0.109} &   \textbf{0.873} &   \underline{4.960} &   \underline{0.209} &   \textbf{0.864} &   \textbf{0.948} &   0.975 \\ 
\arrayrulecolor{gray}\hline

{\bf Monodepth2} (1024 $\times$ 320) &   S & 
  \textbf{0.107} &   \textbf{0.849} &   \textbf{4.764} &   \textbf{0.201} &   \textbf{0.874} &   \textbf{0.953} &   \underline{0.977} \\ 

\arrayrulecolor{black}\hline

UnDeepVO \cite{li2017undeepvo} & MS  &  0.183 & 1.730 & 6.57 & 0.268 & - & - & -\\
Zhan FullNYU \cite{zhanst2018} & D*MS  &  0.135 & 1.132 & 5.585 & 0.229 & 0.820 & 0.933 & 0.971\\
EPC++ \cite{luo2018every} & MS & 0.128 & \underline{0.935} & \underline{5.011} & \underline{0.209} & 0.831 & \underline{0.945} & \textbf{0.979} \\
   \textbf{Monodepth2} w/o pretraining & MS &   
   \underline{0.127} &   1.031 &   5.266 &   0.221 &   \underline{0.836} &   0.943 &   \underline{0.974} \\
\textbf{Monodepth2}& MS &   
   \textbf{0.106} &   \textbf{0.818} &   \textbf{4.750} &   \textbf{0.196} &   \textbf{0.874} &   \textbf{0.957} &   \textbf{0.979} \\

   \arrayrulecolor{gray}\hline
   \textbf{Monodepth2} (1024 $\times$ 320) & MS 
    &   \textbf{0.106} &   \textbf{0.806} &   \textbf{4.630} &   \textbf{0.193} &   \textbf{0.876} &   \textbf{0.958} &   \textbf{0.980}  \\

\arrayrulecolor{black}\hline

  \end{tabular}
  }\hfill
  \raisebox{3pt}{
  \begin{minipage}[c]{0.28\textwidth}
  \caption{\textbf{Quantitative results.} Comparison of our method to existing methods on KITTI 2015 \cite{Geiger2012CVPR} using the Eigen split.  Best results in each category are in \textbf{bold}; second best are \underline{underlined}. \newline
  All results here are presented without post-processing \cite{godard2017unsupervised}; see supplementary Section \ref{sec:post_processing} for improved post-processed results. 
  While our contributions are designed for monocular training, we still gain high accuracy in the stereo-only category. \newline
  We additionally show we can get higher scores at a larger 1024 $\times$ 320 resolution, similar to \cite{pillai2018superdepth} -- see supplementary Section \ref{sec:effect_of_resolution}.
These high resolution numbers are bolded if they beat all other models, including our low-res versions.
  \vspace{8pt}  \newline 
  {\footnotesize
    \textbf{Legend} \newline 
   D -- Depth supervision \newline 
  D* -- Auxiliary depth supervision \newline 
  S -- Self-supervised stereo supervision \newline 
  M -- Self-supervised mono supervision \newline
  \textdagger~ -- Newer results from github. \newline
  + pp -- With post-processing}
  \label{tab:kitti_eigen}}
  \end{minipage}}
  \vspace{-8pt}
\end{table*}

\subsection{Additional Considerations}
Our depth estimation network is based on the general \mbox{U-Net} architecture \cite{ronneberger2015u}, \ie an encoder-decoder network, with skip connections, enabling us to represent both deep abstract features as well as local information.
We use a ResNet18 \cite{he2016deep} as our encoder, which contains 11M parameters, compared to the larger, and slower, DispNet and ResNet50 models used in existing work \cite{godard2017unsupervised}.
Similar to \cite{kuznietsov2017semi,guo2018learning}, we start with weights pretrained on ImageNet~ \cite{russakovsky2015imagenet}, and show that this improves accuracy for our compact model compared to training from scratch (Table \ref{tab:kitti_eigen_ablation}).
Our depth decoder is similar to \cite{godard2017unsupervised}, with sigmoids at the output and ELU nonlinearities \cite{elus} elsewhere.
We convert the sigmoid output $\sigma$ to depth with $D = 1 / ({a\sigma + b})$,
where $a$ and $b$ are chosen to constrain $D$ between 0.1 and 100 units.
We make use of reflection padding, in place of zero padding, in the decoder, returning the value of the closest border pixels in the source image when samples land outside of the image boundaries.
We found that this significantly reduces the border artifacts found in existing approaches, \eg \cite{godard2017unsupervised}.

For pose estimation, we follow \cite{wang2017learning} and predict the rotation using an axis-angle representation, and scale the rotation and translation outputs by $0.01$.
For monocular training, we use a sequence length of three frames, while our pose network is formed from a ResNet18, modified to accept a pair of color images (or six channels) as input and to predict a single 6-DoF relative pose.
We perform horizontal flips and the following training augmentations, with $50\%$ chance: random brightness, contrast, saturation, and hue jitter with respective ranges of $\pm0.2$, $\pm0.2$, $\pm0.2$, and $\pm0.1$. 
Importantly, the color augmentations are only applied to the images which are fed to the networks, not to those used to compute $L_p$. 
All three images fed to the pose and depth networks are augmented with the same parameters.

Our models are implemented in PyTorch~\cite{paszke2017automatic}, trained for 20 epochs using Adam \cite{kingma2014adam}, with a batch size of $12$ and an input/output resolution of $640\times192$ unless otherwise specified.
We use a learning rate of $10^{-4}$ for the first 15 epochs which is then dropped to $10^{-5}$ for the remainder.
This was chosen using a dedicated validation set of $10\%$ of the data.
The smoothness term $\lambda$ is set to $0.001$.
Training takes 8, 12, and 15 hours on a single Titan Xp, for the stereo (S), monocular (M), and monocular plus stereo models (MS).

\section{Experiments}
\vspace{-5pt}
Here, we validate that (1) our reprojection loss helps with occluded pixels compared to existing pixel-averaging, (2) our auto-masking improves results, especially when training on scenes with static cameras, and (3) our multi-scale appearance matching loss improves accuracy. %
We evaluate our models, named {\bf Monodepth2}, on the KITTI 2015 stereo dataset \cite{Geiger2012CVPR}, to allow comparison with previously published monocular methods.

\subsection{KITTI Eigen Split}
We use the data split of Eigen~\etal~\cite{eigen2015predicting}.
Except in ablation experiments, for training which uses monocular sequences (\ie monocular and monocular plus stereo) we follow Zhou~\etal's~\cite{zhou2017unsupervised} pre-processing to remove static frames.
This results in 39,810 monocular triplets for training and 4,424 for validation.
We use the same intrinsics for all images, setting the principal point of the camera to the image center and the focal length to the average of all the focal lengths in KITTI.
For stereo and mixed training (monocular plus stereo), we set the transformation between the two stereo frames to be a pure horizontal translation of fixed length.
During evaluation, we cap depth to 80m per standard practice \cite{godard2017unsupervised}.
For our monocular models, we report results using the per-image median ground truth scaling introduced by \cite{zhou2017unsupervised}.
See also supplementary material Section \ref{sec:single_scale_eval} for results where we apply a single median scaling to the whole test set, instead of scaling each image independently.
For results that use any stereo supervision we do not perform median scaling as scale can be inferred from the known camera baseline during training.

\begin{figure*}[!ht]
  \centering
  \resizebox{\textwidth}{!}{
  \newcommand{\turnheightnew}{0.195\columnwidth}

\centering

\begin{tabular}{@{\hskip 2mm}c@{\hskip 2mm}c@{\hskip 2mm}c@{\hskip 2mm}c@{\hskip 2mm}c@{}}

{\rotatebox{90}{\hspace{4mm}Input}} &
\includegraphics[height=\turnheightnew]{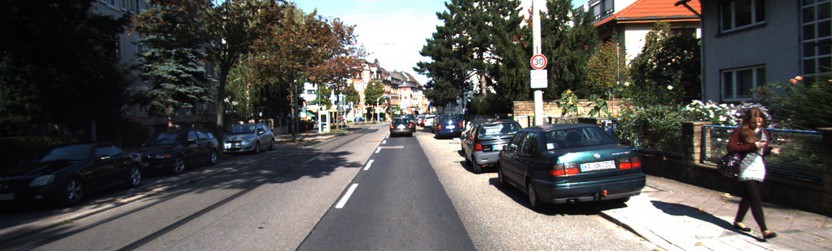} &
\includegraphics[height=\turnheightnew]{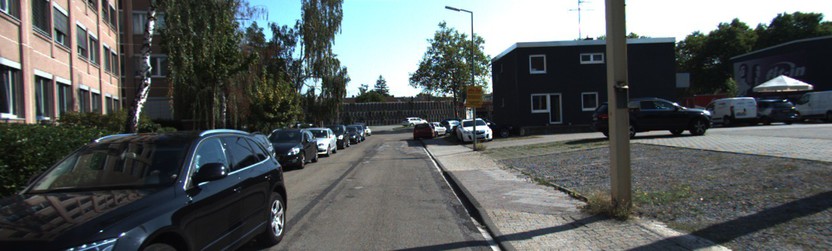} &
\includegraphics[height=\turnheightnew]{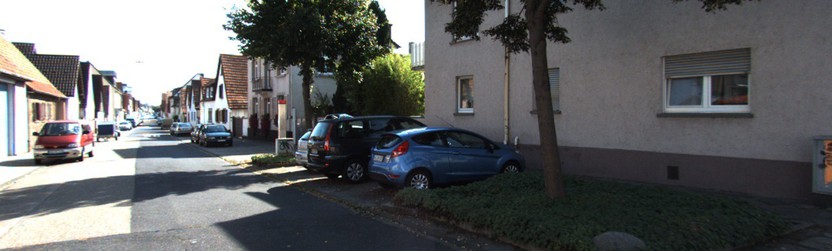} &
\includegraphics[height=\turnheightnew]{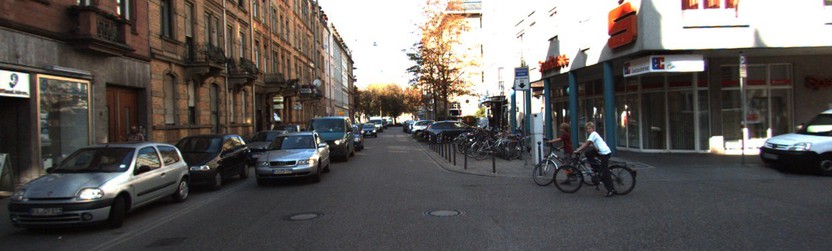}\\

{\rotatebox{90}{\hspace{0mm}\scriptsize
{Monodepth \cite{godard2017unsupervised}}}} &
\includegraphics[height=\turnheightnew]{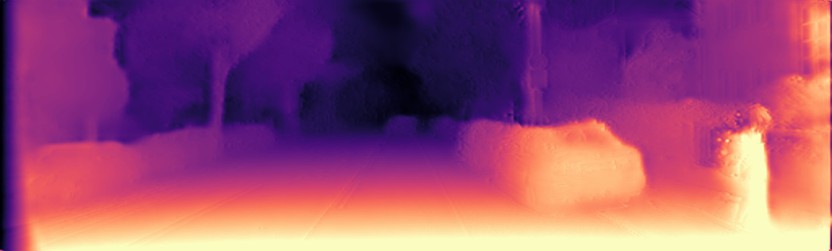} &
\includegraphics[height=\turnheightnew]{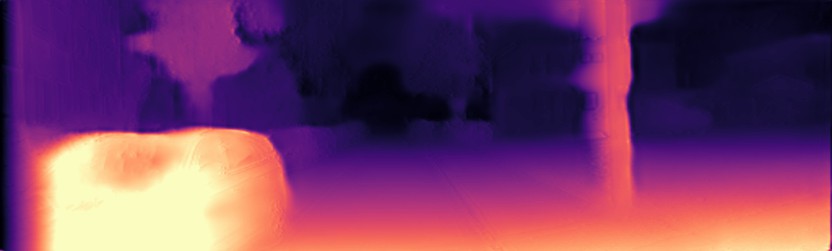} &
\includegraphics[height=\turnheightnew]{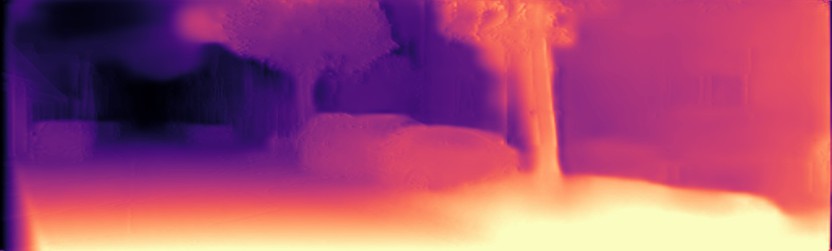} &
\includegraphics[height=\turnheightnew]{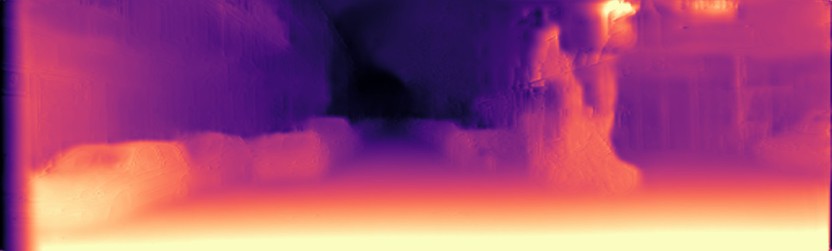} \\

{\rotatebox{90}{\hspace{0mm}\scriptsize
{Zhou \ea~\cite{zhou2017unsupervised}}}} &
\includegraphics[height=\turnheightnew]{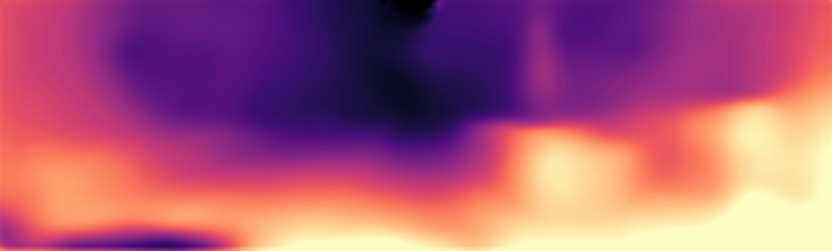} &
\includegraphics[height=\turnheightnew]{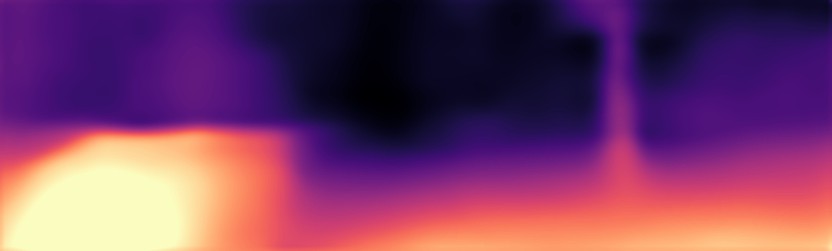} &
\includegraphics[height=\turnheightnew]{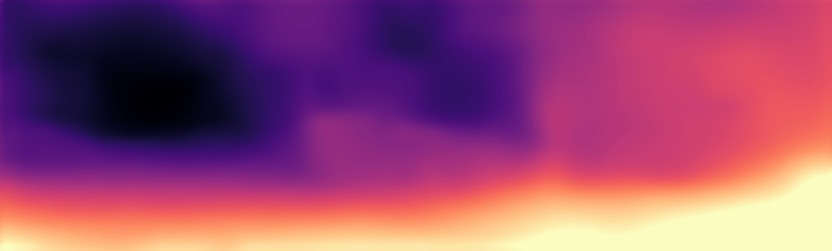} &
\includegraphics[height=\turnheightnew]{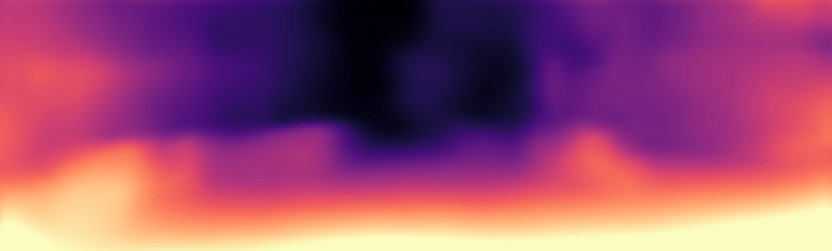} \\

{\rotatebox{90}{\hspace{2mm}\scriptsize
{DDVO~\cite{wang2017learning}}}} &
\includegraphics[height=\turnheightnew]{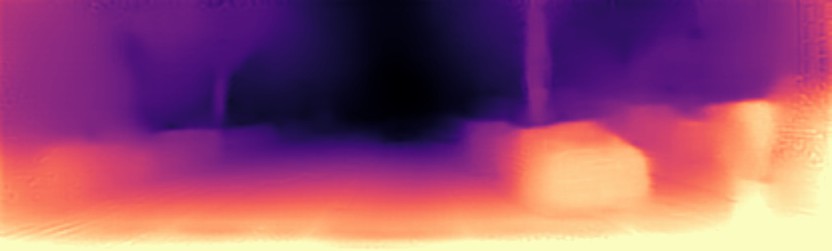} &
\includegraphics[height=\turnheightnew]{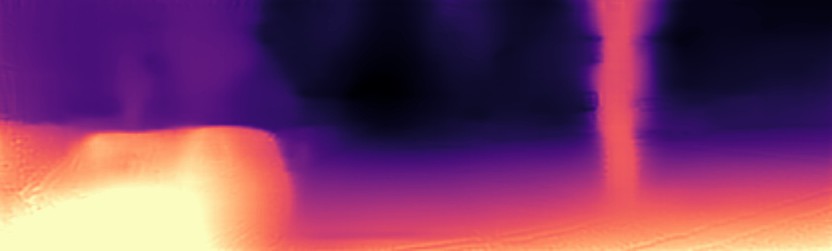} &
\includegraphics[height=\turnheightnew]{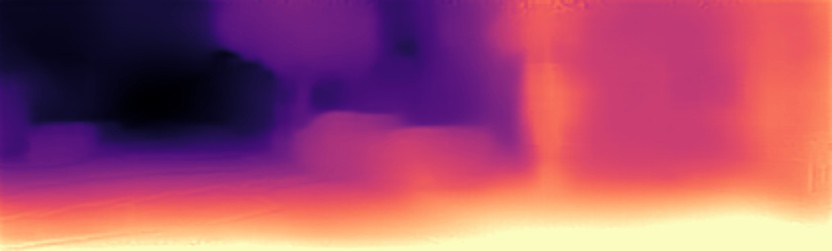} &
\includegraphics[height=\turnheightnew]{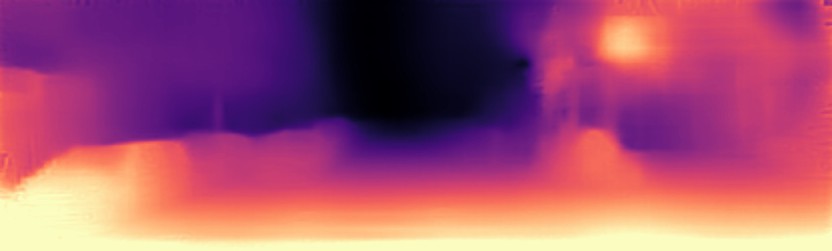} \\

{\rotatebox{90}{\hspace{2mm}\scriptsize
{GeoNet~\cite{geonet2018}}}} &
\includegraphics[height=\turnheightnew]{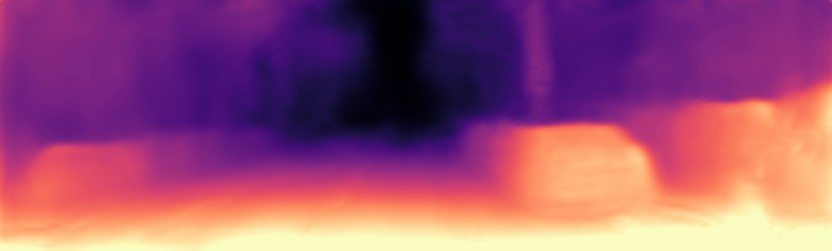} &
\includegraphics[height=\turnheightnew]{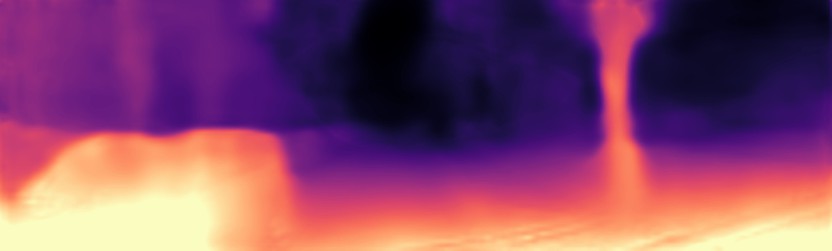} &
\includegraphics[height=\turnheightnew]{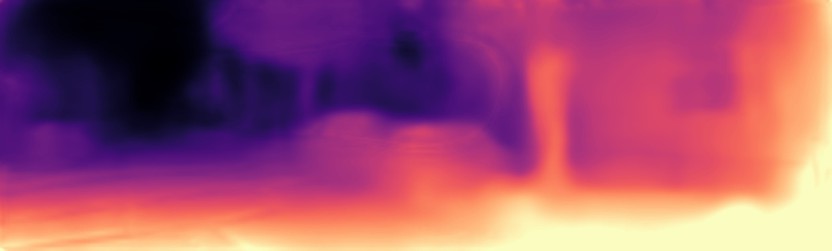} &
\includegraphics[height=\turnheightnew]{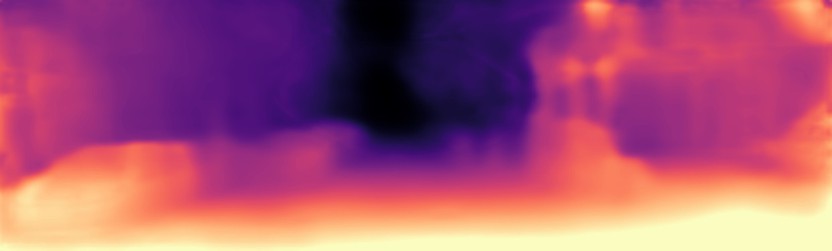} \\

{\rotatebox{90}{\hspace{0mm}\scriptsize {Zhan \ea~\cite{zhanst2018}}}} &
\includegraphics[height=\turnheightnew]{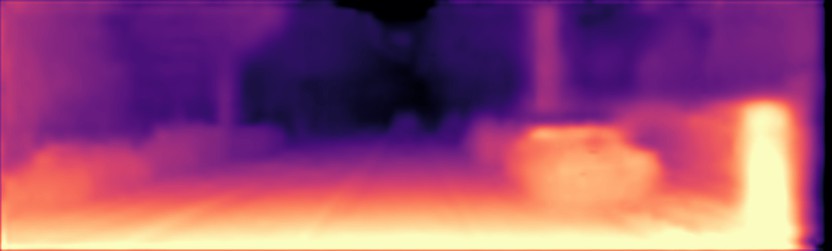} &
\includegraphics[height=\turnheightnew]{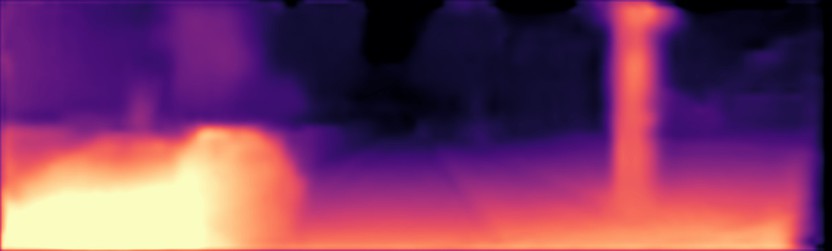} &
\includegraphics[height=\turnheightnew]{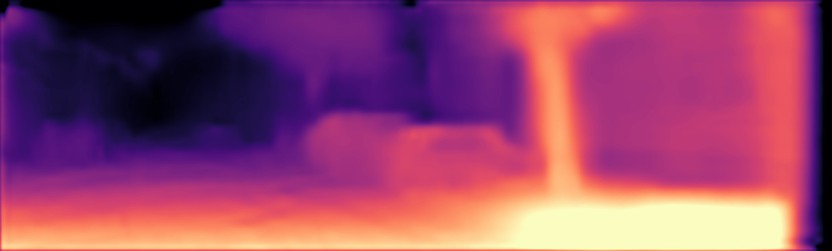} &
\includegraphics[height=\turnheightnew]{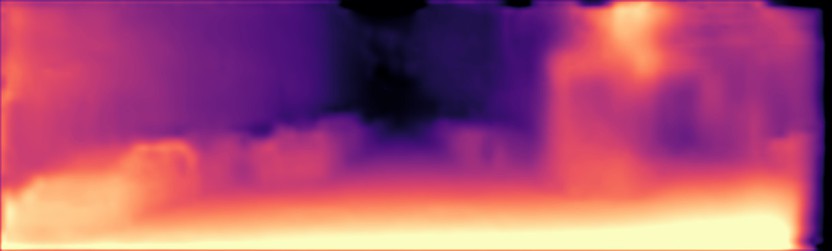} \\

{\rotatebox{90}{\hspace{-1mm}\scriptsize Ranjan \ea~\cite{ranjan2018adversarial}}} &
\includegraphics[height=\turnheightnew]{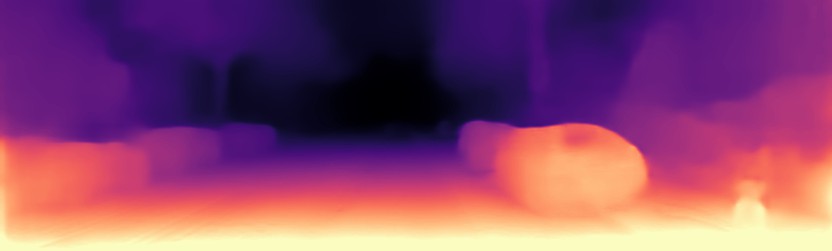} &
\includegraphics[height=\turnheightnew]{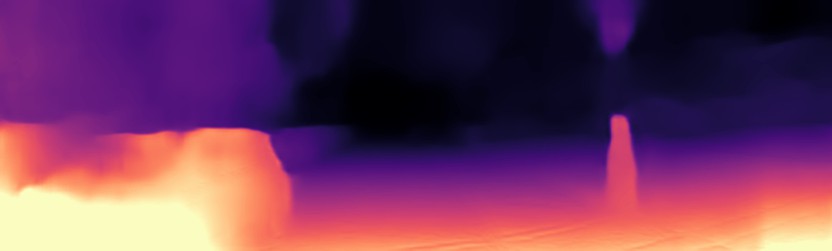} &
\includegraphics[height=\turnheightnew]{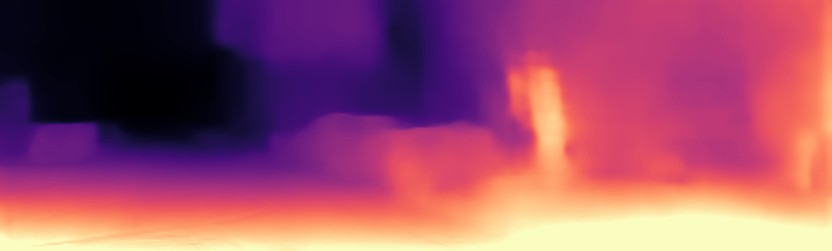} &
\includegraphics[height=\turnheightnew]{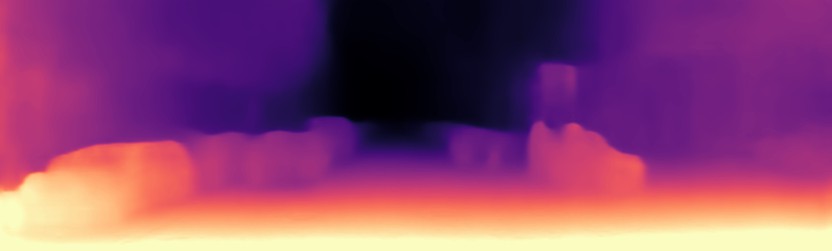} \\

{\rotatebox{90}{\hspace{0mm} \scriptsize 3Net - R50 \cite{luo2018every}}} &
\includegraphics[height=\turnheightnew]{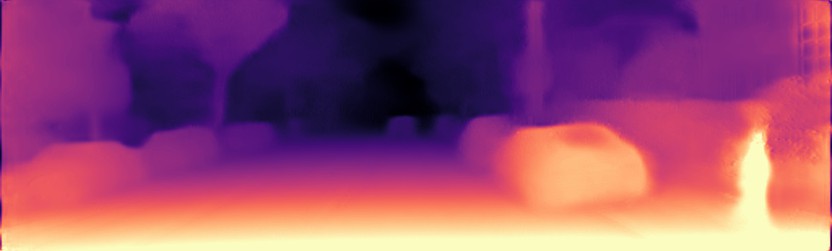} &
\includegraphics[height=\turnheightnew]{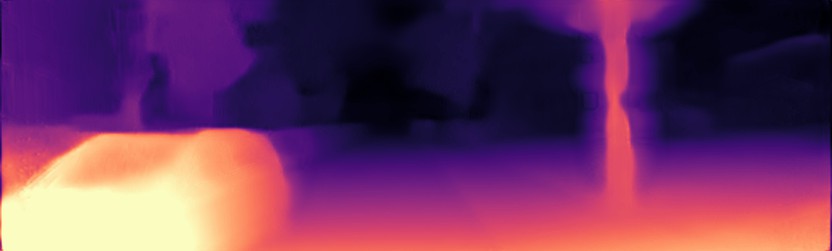} &
\includegraphics[height=\turnheightnew]{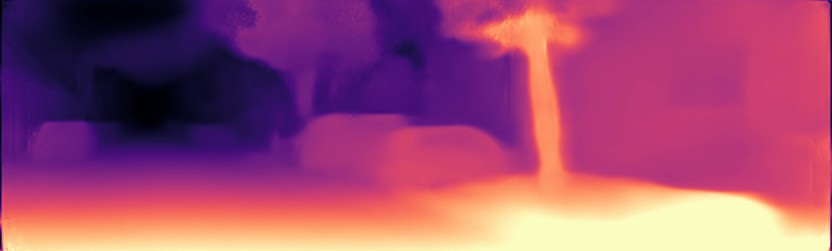} &
\includegraphics[height=\turnheightnew]{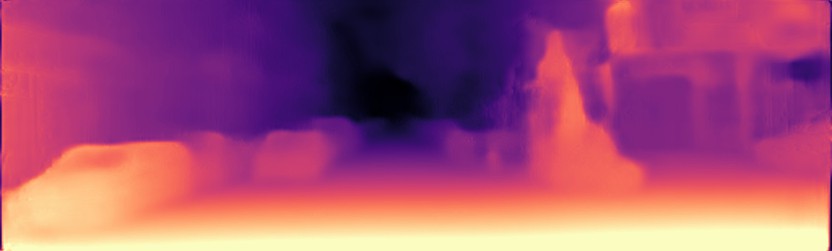} \\ 

{\rotatebox{90}{\hspace{-3mm} \scriptsize EPC++ (MS) \newline \cite{luo2018every}}} &
\includegraphics[height=\turnheightnew]{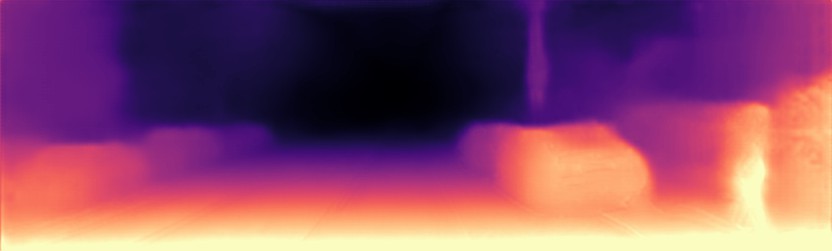} &
\includegraphics[height=\turnheightnew]{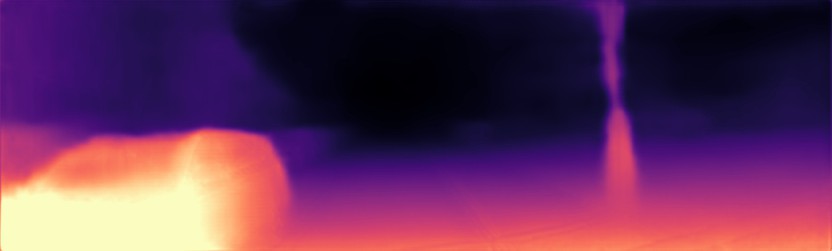} &
\includegraphics[height=\turnheightnew]{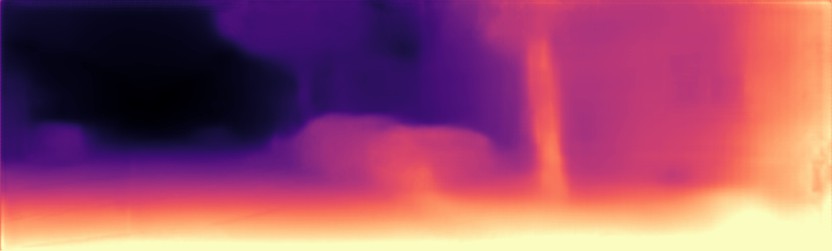} &
\includegraphics[height=\turnheightnew]{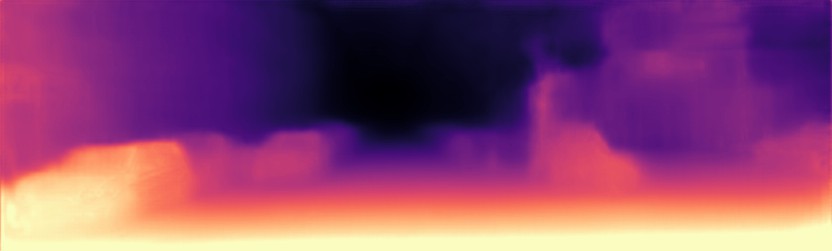} \\ 

{\rotatebox{90}{\scriptsize\hspace{3mm}\textbf{MD2 M}}} &
\includegraphics[height=\turnheightnew]{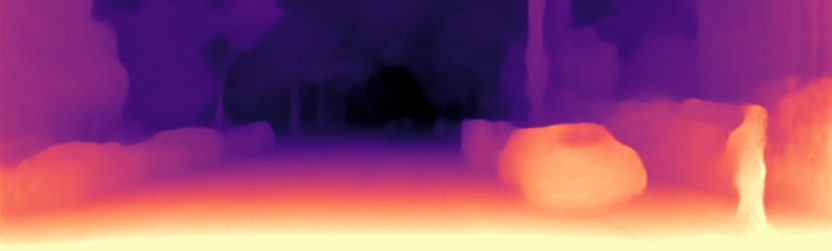} &
\includegraphics[height=\turnheightnew]{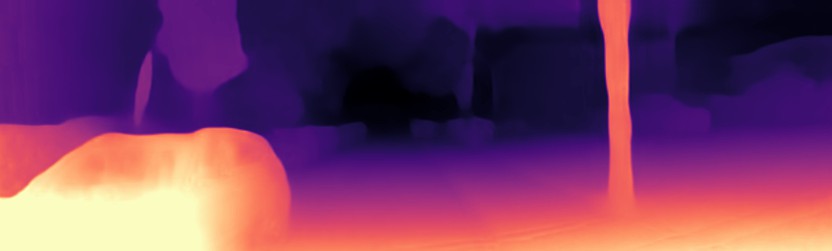} &
\includegraphics[height=\turnheightnew]{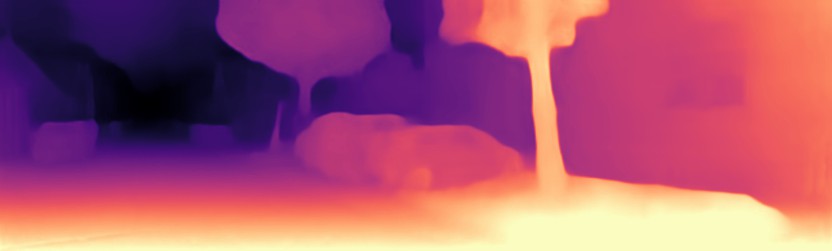} &
\includegraphics[height=\turnheightnew]{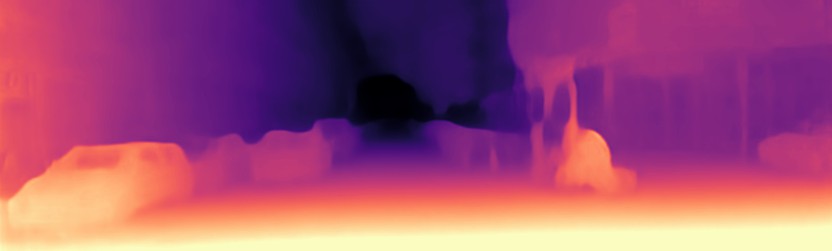} \\

{\rotatebox{90}{\scriptsize \hspace{0mm}\textbf{MD2 M} (no p/t)}} &
\includegraphics[height=\turnheightnew]{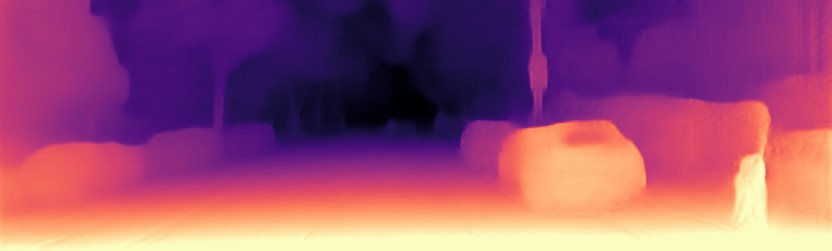} &
\includegraphics[height=\turnheightnew]{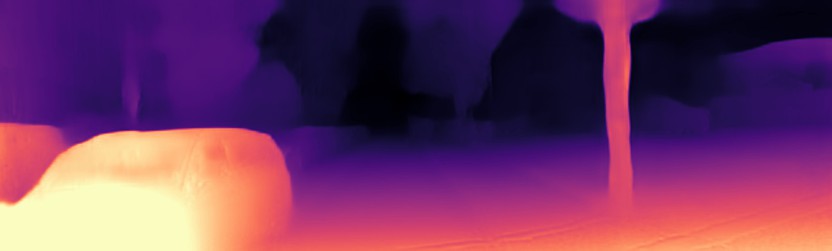} &
\includegraphics[height=\turnheightnew]{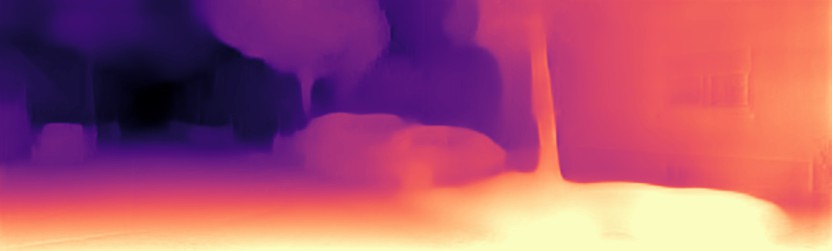} &
\includegraphics[height=\turnheightnew]{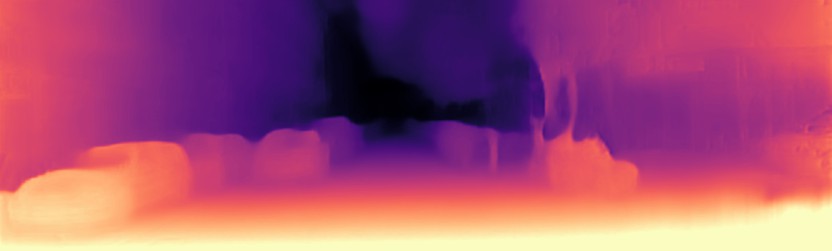} \\

{\rotatebox{90}{\scriptsize\hspace{3mm}\textbf{MD2 S}}} &
\includegraphics[height=\turnheightnew]{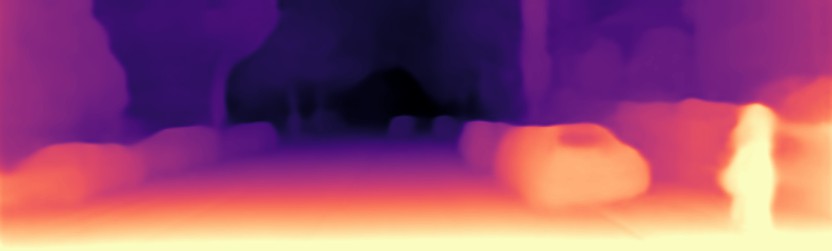} &
\includegraphics[height=\turnheightnew]{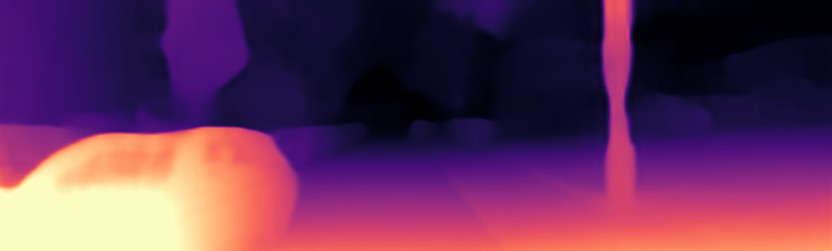} &
\includegraphics[height=\turnheightnew]{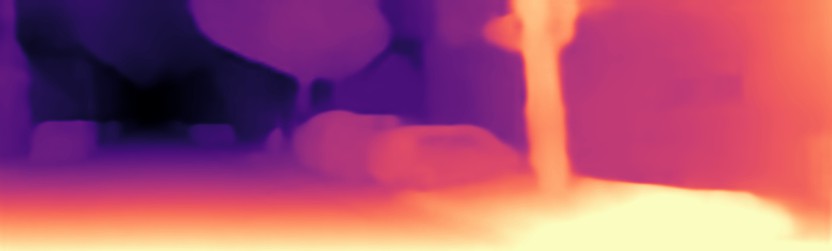} &
\includegraphics[height=\turnheightnew]{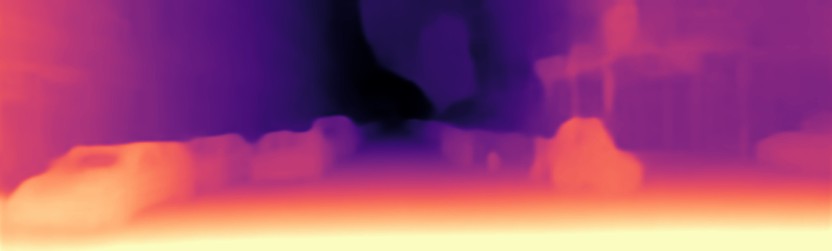} \\

{\rotatebox{90}{\scriptsize\hspace{3mm}\textbf{MD2 MS}}} &
\includegraphics[height=\turnheightnew]{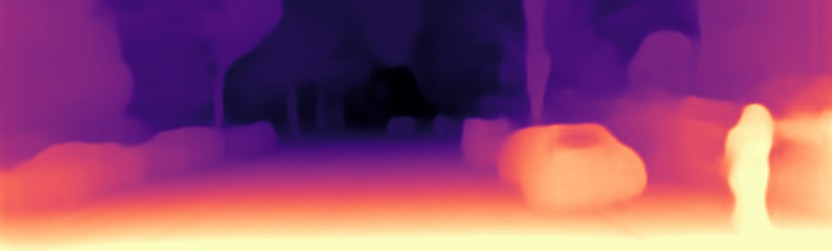} &
\includegraphics[height=\turnheightnew]{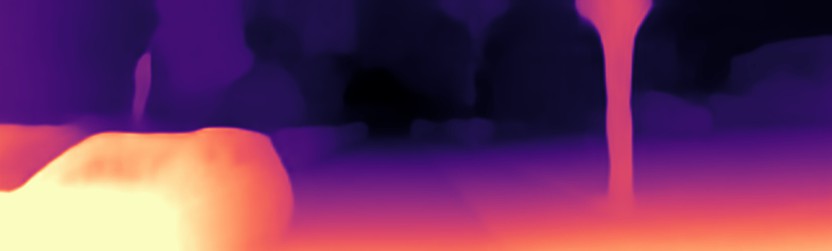} &
\includegraphics[height=\turnheightnew]{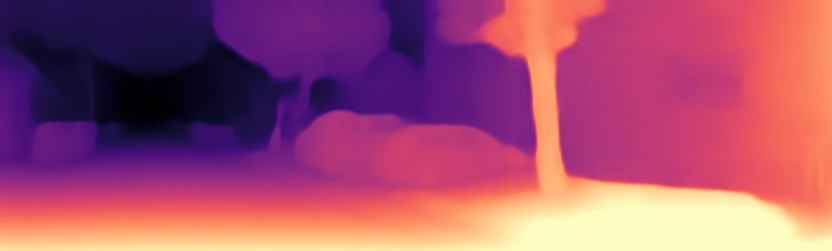} &
\includegraphics[height=\turnheightnew]{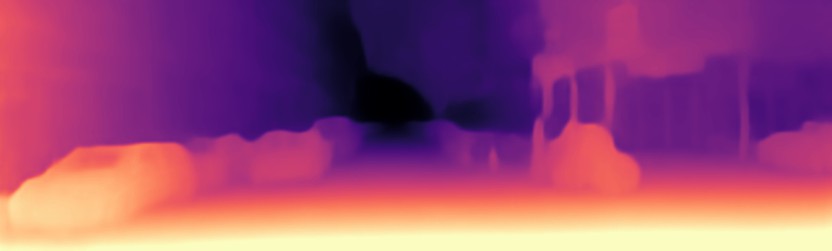} \\

\end{tabular}
 }
  \caption{\textbf{Qualitative results on the KITTI Eigen split.} Our models (MD2) in the last four rows produce the sharpest depth maps, which are reflected in the superior quantitative results in Table~\ref{tab:kitti_eigen}. Additional results can be seen in the supplementary materiale Section \ref{sec:additional_results}.}
  \label{fig:kitti_eigen_qual}
\end{figure*}

We compare the results of several variants of our model, trained with different types of self-supervision: monocular video only (M), stereo only (S), and both (MS).
The results in Table~\ref{tab:kitti_eigen} show that our monocular method outperforms all existing state-of-the-art self-supervised approaches.
We also outperform recent methods (\cite{luo2018every, ranjan2018adversarial}) that explicitly compute optical flow as well as motion masks.
Qualitative results can be seen in Fig.~\ref{fig:kitti_eigen_qual} and supplementary Section \ref{sec:additional_results}.
However, as with all image reconstruction based approaches to depth estimation, our model breaks when the scene contains objects that violate the Lambertian assumptions of our appearance loss (Fig. \ref{fig:failure}).

As expected, the combination of M and S training data increases accuracy, which is especially noticeable on metrics that are sensitive to large depth errors \eg RMSE.
Despite our contributions being designed around monocular training, we find that the in the stereo-only case we still perform well.
We achieve high accuracy despite using a lower resolution than \cite{pillai2018superdepth}'s $1024\times384$, with substantially less training time (20 vs.~200 epochs) and no use of post-processing.

\newcolumntype{P}[1]{>{\centering\arraybackslash}p{#1}}

\definecolor{light-gray}{gray}{0.85}
\newcommand{\hlinegray}{\arrayrulecolor{light-gray}\hline\arrayrulecolor{black}}
\newcommand{\clinegrayone}{\arrayrulecolor{light-gray}\cline{2-14}\arrayrulecolor{black}}
\newcommand{\clineblackone}{\arrayrulecolor{black}\cline{2-14}\arrayrulecolor{black}}

\setlength\tabcolsep{4pt} %

\begin{table*}[t!]
  \centering
  \resizebox{\textwidth}{!}
{
    \footnotesize
    \begin{tabular}{|l|l||c|c|c||c|c||c|c|c|c|c|c|c|}
      \hline
      & & 
      \begin{tabular}{@{}c@{}}Auto- \\ masking\end{tabular} & 
      \begin{tabular}{@{}c@{}}Min. \\ reproj.\end{tabular} & 
      \begin{tabular}{@{}c@{}}Full-res \\ multi-scale\end{tabular} &
      
      Pretrained &  
      \begin{tabular}{@{}c@{}}Full Eigen \\ split\end{tabular} &
      \cellcolor{col1}Abs Rel & \cellcolor{col1}Sq Rel & \cellcolor{col1}RMSE  &
      \cellcolor{col1}\begin{tabular}{@{}c@{}}RMSE \\ log\end{tabular} & 
      \cellcolor{col2}$\delta < $1.25 & \cellcolor{col2}$\delta < $1.25$^{2}$ & \cellcolor{col2}$\delta <$ $1.25^{3}$ \\
      
      \hline

      (a) &  Baseline    &   &   &  & \checkmark &         &  
          0.140 &   1.610 &   5.512 &   0.223 &   0.852 &   0.946 &   0.973 \\  %
          \clinegrayone
  
        & Baseline + min reproj. &  & \checkmark &     & \checkmark &  & 
            0.122  &   1.081  &   5.116  &   0.199  &   0.866  &   0.957  &   0.980  \\
        \clinegrayone

            & Baseline + automasking & \checkmark & &  &  & 
        &   0.124  &   0.936  &   5.010  &   0.206  &   0.858  &   0.952  &   0.977  \\
       \clinegrayone

          & Baseline + full-res m.s.  &  & & \checkmark   & \checkmark &  & 
            0.124  &  1.170  &  5.249  &   0.203  & 0.865  & 0.953  &   0.978  \\
            \clinegrayone

    & Monodepth2 w/o min~reprojection &
      \checkmark &  & \checkmark &  \checkmark &  &   
      0.117 &   0.878 &   4.846 &   0.196 &   0.870 &   0.957 &   0.980 \\ %
      \clinegrayone
      
      & Monodepth2 w/o auto-masking &  & \checkmark & \checkmark & \checkmark &  & 
      0.120 &   1.097 &   5.074 &   0.197 &   0.872 &   0.956 &   0.979 \\  %
      \clinegrayone

      & Monodepth2 w/o full-res m.s. & \checkmark & \checkmark &  &  \checkmark &   & 
      0.117 &   {\bf 0.866} &   4.864 &   0.196 &   0.871 &   0.957 &   {\bf 0.981}  \\  %
      \clinegrayone

      & Monodepth2 with \cite{zhou2017unsupervised}'s mask &  & \checkmark & \checkmark & \checkmark &    &
       0.123 &   1.177 &   5.210 &   0.200 &   0.869 &   0.955 &   0.978 \\  %
      \clinegrayone

      & Monodepth2 smaller (416 $\times$ 128)             &  \checkmark &  \checkmark & \checkmark & \checkmark &  & 
         0.128 &   1.087 &   5.171 &   0.204 &   0.855 &   0.953 &   0.978 \\
      \clinegrayone

      & \textbf{Monodepth2} (full)              &  \checkmark &  \checkmark & \checkmark & \checkmark &  & 
        {\bf 0.115} &   0.903 &   {\bf 4.863} &   {\bf 0.193} &   {\bf 0.877} &   {\bf 0.959} &   {\bf 0.981} \\ %

      \hline %

      (b) & Baseline w/o pt    &  & &  &  &  & 
        0.150 &   1.585 &   5.671 &   0.234 &   0.827 &   0.938 &   0.971 \\ %
     
      \clinegrayone

      & \textbf{Monodepth2}  w/o pt              &  \checkmark &  \checkmark & \checkmark &  &  & 
        {\bf 0.132} &   {\bf 1.044} &   {\bf 5.142} &   {\bf 0.210} &   {\bf 0.845} &   {\bf 0.948} &   {\bf 0.977} \\  %

      \hline %
    
     (c)   & Baseline (full Eigen dataset)  &   &   &  & \checkmark &    \checkmark     & 
     0.146 &   1.876 &   5.666 &   0.230 &   0.848 &   0.945 &   0.972 \\ %
     
     \clinegrayone
     
     & \textbf{Monodepth2} (full Eigen dataset) &  \checkmark &  \checkmark & \checkmark & \checkmark & \checkmark & 
     {\bf 0.116} &   {\bf 0.918} &   {\bf 4.872} &   {\bf 0.193} &   {\bf 0.874} &   {\bf 0.959} &   {\bf 0.981} \\ 
      
      \hline %
      
      \end{tabular}
  }
    \vspace{0pt}
      \caption{
      \textbf{Ablation.} 
      Results for different variants of our model (\textbf{Monodepth2}) with monocular training on KITTI 2015 \cite{Geiger2012CVPR} using the Eigen split.
      \textbf{(a)} The baseline model, with none of our contributions, performs poorly.
      The addition of our minimum reprojection, auto-masking and full-res multi-scale components, significantly improves performance.
      \textbf{(b)} Even without ImageNet pretrained weights, our much simpler model brings large improvements above the baseline -- see also Table~\ref{tab:kitti_eigen}.  
      \textbf{(c)} If we train with the full Eigen dataset (instead of the subset introduced for monocular training by \cite{zhou2017unsupervised}) our improvement over the baseline increases. }
      \vspace{-2pt}
\label{tab:kitti_eigen_ablation}
\end{table*}

\subsubsection{KITTI Ablation Study} 
\vspace{-5pt}
To better understand how the components of our model contribute to the overall performance in monocular training, in Table~\ref{tab:kitti_eigen_ablation}(a) we perform an ablation study by changing various components of our model. %
We see that the baseline model, without any of our contributions, performs the worst.
When combined together, all our components lead to a significant improvement (\textbf{Monodepth2}~(full)).
More experiments turning parts of our full model off in turn are shown in supplementary material Section \ref{sec:additional_ablation}.

\vspace{5pt}
\noindent{}{\bf Benefits of auto-masking}
The full Eigen \cite{eigen2015predicting} KITTI split contains several sequences where the camera does not move between frames \eg where the data capture car was stopped at traffic lights.
These `no camera motion' sequences can cause problems for self-supervised monocular training, and as a result, they are typically excluded at training time using expensive to compute optical flow~\cite{zhou2017unsupervised}. 
We report monocular results trained on the full Eigen data split in Table~\ref{tab:kitti_eigen_ablation}(c), \ie without removing frames.
The baseline model trained on the full KITTI split performs worse than our full model. 
Further, in Table~\ref{tab:kitti_eigen_ablation}(a), we replace our auto-masking loss with a re-implementation of the predictive mask from \cite{zhou2017unsupervised}. 
We find that this ablated model is worse than using no masking at all, while our auto-masking improves results in all cases. 
We see an example of how auto-masking works in practice in Fig.~\ref{fig:automasking}.

\vspace{5pt}
\noindent{}{\bf Effect of ImageNet pretraining}
We follow previous work \cite{girshick2014rich, kuznietsov2017semi,guo2018learning} in initializing our encoders with weights pretrained on ImageNet \cite{russakovsky2015imagenet}.
While some other monocular depth prediction works have elected not to use ImageNet pretraining, we show in Table~\ref{tab:kitti_eigen} that even without pretraining, we still achieve state-of-the-art results.
We train these `w/o pretraining' models for 30 epochs to ensure convergence.
Table~\ref{tab:kitti_eigen_ablation} shows the benefit our contributions bring both to pretrained networks and those trained from scratch; see supplementary material Section \ref{sec:additional_ablation} for more ablations.

\begin{figure}
  \centering
  \vspace{-1pt}
  \resizebox{\columnwidth}{!}{
  \newcommand{\turnheightnew}{0.195\columnwidth}

\centering

\begin{tabular}{@{\hskip 1mm}c@{\hskip 1mm}c@{\hskip 1mm}c@{\hskip 1mm}c@{\hskip 1mm}c@{}}

\Large{Input} & \Large{Zhou~\ea\cite{zhou2017unsupervised}}  & \Large{DDVO~\cite{wang2017learning}}  & \Large{Monodepth2 (M)} & \Large{Ground truth} \\

\includegraphics[height=\turnheightnew]{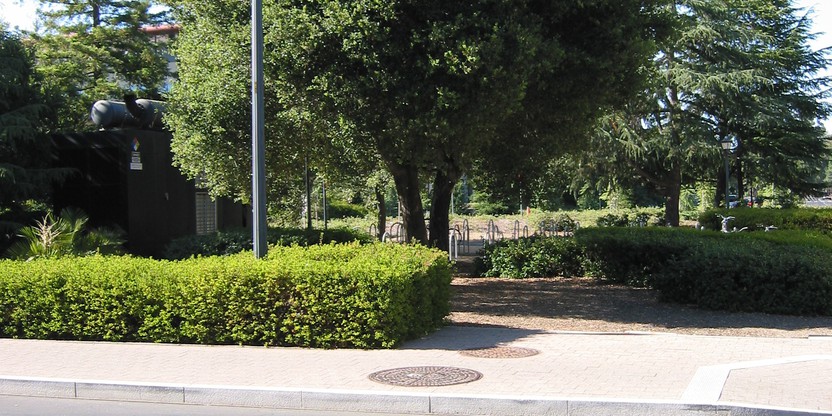} &
\includegraphics[height=\turnheightnew]{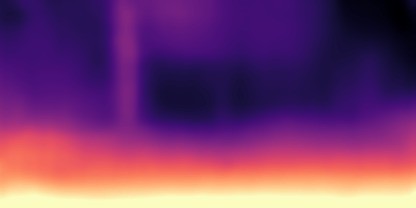} &
\includegraphics[height=\turnheightnew]{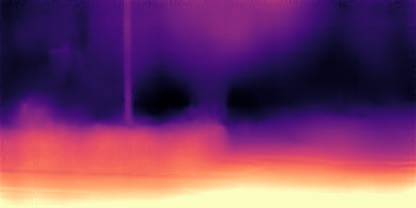} &
\includegraphics[height=\turnheightnew]{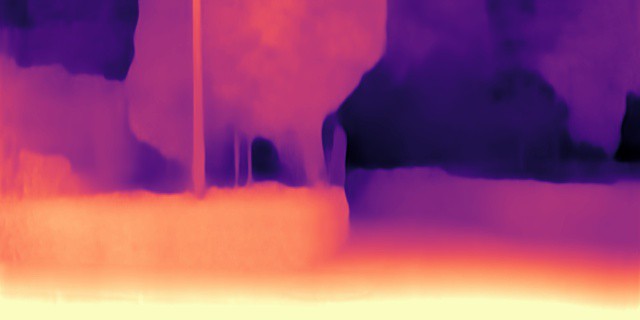} &
\includegraphics[height=\turnheightnew]{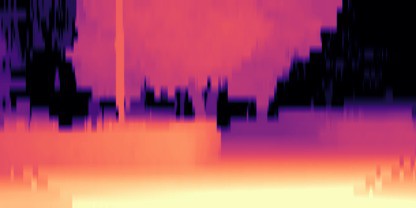}\\

\includegraphics[height=\turnheightnew]{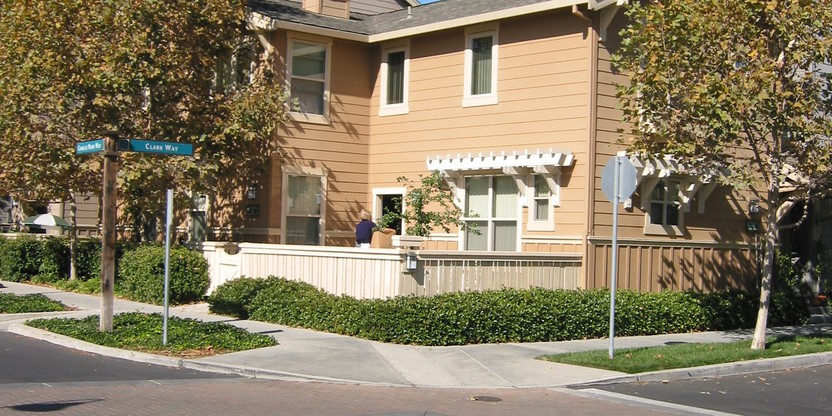} &
\includegraphics[height=\turnheightnew]{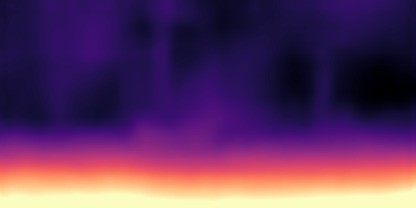} &
\includegraphics[height=\turnheightnew]{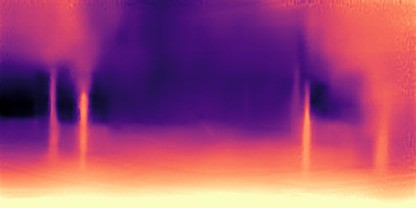} &
\includegraphics[height=\turnheightnew]{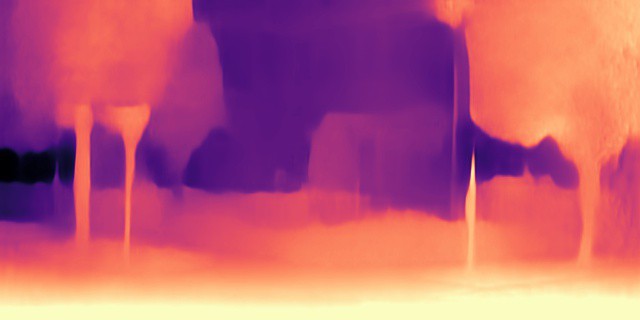} &
\includegraphics[height=\turnheightnew]{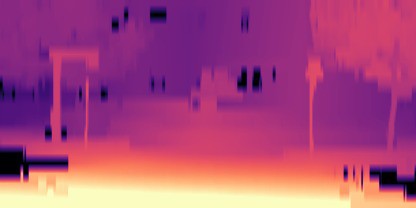}\\

\end{tabular}
 }
  \vspace{-2pt}
  \caption{\textbf{Qualitative Make3D results.} All methods were trained on KITTI using monocular supervision.}
  \label{fig:make3d_results}
  \vspace{-8pt}
\end{figure}

\newcommand{\shiftleft}[2]{\makebox[0pt][r]{\makebox[#1][l]{#2}}}

\newcommand{\imlabel}[2]{\includegraphics[width=0.49\columnwidth]{#1}%
\raisebox{25pt}{\shiftleft{75pt}{\makebox[-2pt][r]{\footnotesize #2}}}}

\subsection{Additional Datasets}

\noindent{}{\bf KITTI Odometry}
In Section \ref{sec:odometry_eval} of the supplementary material we show odometry evaluation on KITTI.
While our focus is better depth estimation, our pose network performs on par with competing methods.
Competing methods typically feed more frames to their pose network which may improve their ability to generalize.

\vspace{5pt}
\noindent{}{\bf KITTI Depth Prediction Benchmark}
\label{ref:kitti_benchmark}
We also perform experiments on the recently introduced KITTI Depth Prediction Evaluation dataset \cite{uhrig2017sparse}, which features more accurate ground truth depth, addressing quality issues with the standard split.
We train models using this new benchmark split, and evaluate it using the online server \cite{kittidepthserver}, and provide results in supplementary Section \ref{sec:kitt_online_bench}.
Additionally, 93\% of the Eigen split test frames have higher quality ground truth depths provided by \cite{uhrig2017sparse}. 
Like \cite{aleotti2018generative}, we use these instead of the reprojected LIDAR scans to compare our method against several existing baseline algorithms, still showing superior performance.

\renewcommand{\imlabel}[2]{\includegraphics[width=0.49\columnwidth]{#1}%
\raisebox{25pt}{\shiftleft{52pt}{\makebox[-2pt][r]{\footnotesize #2}}}}

\begin{figure}
  \centering
  \resizebox{1.0\columnwidth}{!}{
  \newcommand{\turnheightnew}{0.195\columnwidth}
        \centering
        \begin{tabular}{@{\hskip 0mm}c@{\hskip 0mm}c}
            \imlabel{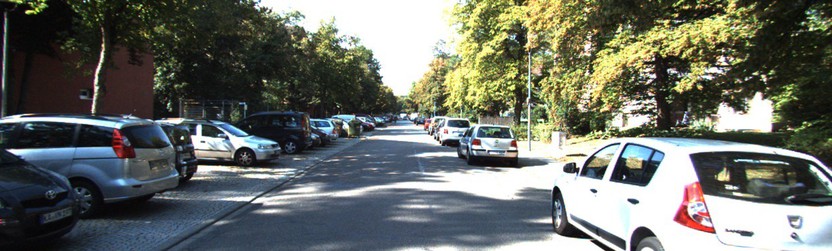}{\textcolor{black}{}} &
            \hspace{0pt} \imlabel{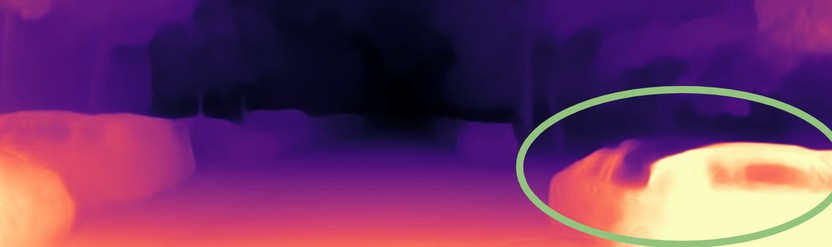}{\textcolor{white}{\textbf{Monodepth2} (M)}} \\
            \imlabel{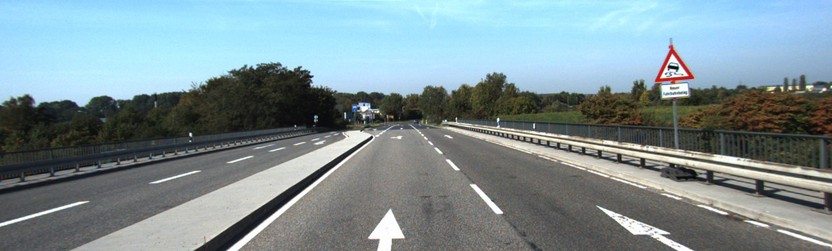}{\textcolor{black}{}} &
            \hspace{0pt} \imlabel{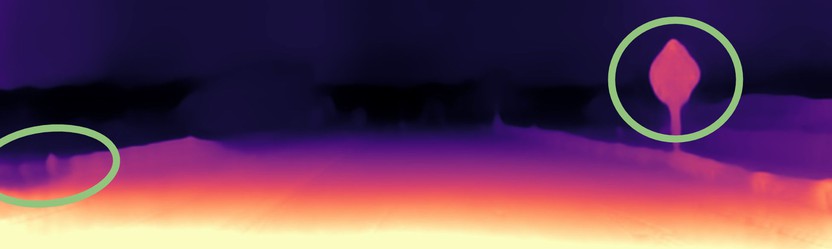}{\textcolor{white}{\textbf{Monodepth2} (M)}} 
        \end{tabular}
    }
  \caption{\textbf{Failure cases.} \textbf{Top:} Our self-supervised loss fails to learn good depths for distorted, reflective and color-saturated regions. \textbf{Bottom:} We can fail to accurately delineate objects where boundaries are ambiguous (left) or shapes are intricate (right). }
    \vspace{-4pt}
  \label{fig:failure}
\end{figure}

\begin{table}
    \centering
    \resizebox{\linewidth}{!}{
      \begin{tabular}{|l|c||c|c|c|c|}
      \hline
       & Type & Abs Rel & Sq Rel  & RMSE & $\text{log}_{10}$ \\
      \hline
      Karsch \cite{karsch2014depth} & D & 0.428 & 5.079 & 8.389 & 0.149 \\
      Liu \cite{liu2014discrete}& D & 0.475 & 6.562 & 10.05 & 0.165 \\
      Laina \cite{laina2016deeper}& D & {\bf 0.204} & {\bf 1.840} & {\bf 5.683} & {\bf 0.084} \\ \hline
      Monodepth \cite{godard2017unsupervised} & S & 0.544 & 10.94 & 11.760 & 0.193 \\
      Zhou \cite{zhou2017unsupervised} & M &  0.383 & 5.321 & 10.470 & 0.478 \\
      DDVO \cite{wang2017learning} & M & 0.387 & 4.720 & 8.090 & 0.204 \\
        \textbf{Monodepth2} & M & \textbf{0.322} & \textbf{3.589} & \textbf{7.417} & \textbf{0.163}  \\
    \hline
    \textbf{Monodepth2} & MS & 0.374 & 3.792 & 8.238 & 0.201 \\ %
    
      \hline
      \end{tabular}
    }
    \vspace{2.0pt}
    \caption{{\bf Make3D results.} All M results benefit from median scaling, while MS uses the unmodified network prediction.}
    \label{tab:make3d}
    \vspace{-8pt}
\end{table}

\vspace{5pt}
\noindent{}{\bf Make3D}
In Table~\ref{tab:make3d} we report performance on the Make3D dataset~\cite{saxena2009make3d} using our models trained on KITTI.
We outperform all methods that do not use depth supervision, with the evaluation criteria from \cite{godard2017unsupervised}.
However, caution should be taken with Make3D, as its ground truth depth and input images are not well aligned, causing potential evaluation issues. 
We evaluate on a center crop of $2\times1$ ratio, and apply median scaling for our M model.
Qualitative results can be seen in Fig.~\ref{fig:make3d_results} and in supplementary Section \ref{sec:additional_results}.

\section{Conclusion}
\vspace{-4pt}
We have presented a versatile model for self-supervised monocular depth estimation, achieving state-of-the-art depth predictions.
We introduced three contributions:
(i) a minimum reprojection loss, computed for each pixel, to deal with occlusions between frames in monocular video, 
(ii) an auto-masking loss to ignore confusing, stationary pixels, and
(iii) a full-resolution multi-scale sampling method.
We showed how together they give a simple and efficient model for depth estimation, which can be trained with monocular video data, stereo data, or mixed monocular and stereo data.

\vspace{4pt}
\noindent{}\textbf{Acknowledgements} Thanks to the authors who shared their results, and Peter Hedman, Daniyar Turmukhambetov, and Aron Monszpart for their helpful discussions.

\bibliographystyle{ieee_fullname}
{\small
\bibliography{main}
}

\clearpage

\begin{appendices}

\paragraph{Note on arXiv versions}
In an earlier pre-print of this paper, \emph{1806.01260v1}, we included a shared encoder for pose and depth. While this reduced the number of training parameters, we have since found that we can gain even higher results with a separate ResNet pose encoder which accepts a stack of two frames as input (see ablation study in Section~\ref{sec:pose_encoder_comparison}).
Since \emph{v1}, we have also introduced \emph{auto-masking} to help the model ignore pixels that violate our motion assumptions.

\section{Odometry Evaluation}
\label{sec:odometry_eval}
In Table \ref{tab:odom} we evaluate our pose estimation network following the protocol in \cite{zhou2017unsupervised}. 
We trained our models on sequences 0-8 from the KITTI odometry split and tested on sequences 9 and 10. 
As in \cite{zhou2017unsupervised}, the absolute trajectory error is then averaged over all overlapping five-frame snippets in the test sequences. 
Here, unlike \cite{zhou2017unsupervised} and others who use \emph{custom} models for the odometry task, we use the \emph{same} architecture for this task as our other results, and simply train it again from scratch on these new sequences.

Baselines such as \cite{zhou2017unsupervised} use a pose network which predicts transformations between sets of five frames simultaneously.
Our pose network only takes two frames as input, and ouputs a single transformation between that pair of frames.
In order to evaluate our two-frame model on the five-frame test sequences, we make separate predictions for each of the four frame-to-frame transformation for each set of five frames and combine them to form local trajectories.
For completeness we repeat the same process with \cite{zhou2017unsupervised} predicted poses, which we denote as `Zhou$^*$'. 
As we can see in Table \ref{tab:odom}, our frame-to-frame poses come close to the accuracy of methods trained on blocks of five frames at a time. 

\begin{table}[!ht]
\centering
\resizebox{1.0\columnwidth}{!}{
\begin{tabular}{l|c|c|c|}
\cline{2-4}
                           & \textbf{Sequence 09}     & \textbf{Sequence 10} & \textbf{\# frames} \\ \hline
\multicolumn{1}{|l|}{ORB-Slam \cite{mur2015orb}} & 0.014$\pm$0.008 & 0.012$\pm$0.011 & -  \\ \hline \hline
\multicolumn{1}{|l|}{DDVO \cite{wang2017learning}} & 0.045$\pm$0.108 & 0.033$\pm$0.074 & 3 \\ \hline
\multicolumn{1}{|l|}{Zhou* \cite{zhou2017unsupervised}} & 0.050$\pm$0.039 & 0.034$\pm$0.028 & 5 $\to$ 2\\ \hline
\multicolumn{1}{|l|}{Zhou \cite{zhou2017unsupervised}} & 0.021$\pm$0.017 & 0.020$\pm$0.015 & 5\\ \hline
\multicolumn{1}{|l|}{Zhou \cite{zhou2017unsupervised}\textdagger} & 0.016$\pm$0.009 & 0.013$\pm$0.009 & 5 \\ \hline
\multicolumn{1}{|l|}{Mahjourian \cite{mahjourian2018unsupervised}} & 0.013$\pm$0.010 & 0.012$\pm$0.011 & 3 \\ \hline
\multicolumn{1}{|l|}{GeoNet \cite{geonet2018}} & 0.012$\pm$0.007 & 0.012$\pm$0.009 & 5 \\ \hline 
\multicolumn{1}{|l|}{EPC++ M \cite{luo2018every}} & 0.013$\pm$0.007 & \textbf{0.012}$\pm$\textbf{0.008} & 3\\ \hline
\multicolumn{1}{|l|}{Ranjan \cite{ranjan2018adversarial}} & \textbf{0.012}$\pm$0.007 & \textbf{0.012}$\pm$\textbf{0.008} & 5\\ \hline
\multicolumn{1}{|l|}{EPC++ MS \cite{luo2018every}} & \textbf{0.012}$\pm$\textbf{0.006} & \textbf{0.012}$\pm$\textbf{0.008} & 3\\ \hline \hline
\multicolumn{1}{|l|}{\textbf{Monodepth2 M}*} & 0.017$\pm$0.008 & 0.015$\pm$0.010 & 2 \\ \hline
\multicolumn{1}{|l|}{\textbf{Monodepth2 MS}*} & 0.017$\pm$0.008 & 0.015$\pm$0.010 & 2  \\ \hline\hline
\multicolumn{1}{|l|}{\textbf{Monodepth2 M} w/o pretraining*}  & 0.018$\pm$0.010 & 0.015$\pm$0.010 & 2  \\ \hline
\multicolumn{1}{|l|}{\textbf{Monodepth2 MS} w/o pretraining*} &  0.018$\pm$0.009 & 0.015$\pm$0.010 &  2  \\ \hline

\end{tabular}
}
\vspace{1pt}
\caption{
\textbf{Odometry results on the KITTI \cite{Geiger2012CVPR} odometry dataset.} Results show the average absolute trajectory error, and standard deviation, in meters.\vspace{4pt} \newline{\footnotesize \textdagger~-- newer results from the respective online implementations.  \newline  *~-- evaluation on trajectories made from pairwise predictions -- see text for details. \newline `\# frames' is the number of input frames used for pose prediction. To evaluate our method we chain integrate the poses from four pairs to make five frames for evaluation.}}
\label{tab:odom}
\end{table}

\begin{table}
  \centering
  \resizebox{1.0\columnwidth}{!}{
  \begin{tabular}[t]{l}

\begin{tabular}[t]{|l|l|l|l|l|l|l|}
\hline
\multicolumn{7}{|l|}{\textbf{Depth Decoder}} \\
\hline
\textbf{layer} & \textbf{k} & \textbf{s} & \textbf{chns} & \textbf{res} & \textbf{input}   & \textbf{activation}    \\ \hline
upconv5       & 3      & 1      & 256      & 32    & econv5                     & ELU \cite{elus} \\
iconv5        & 3      & 1      & 256      & 16    & $\uparrow$upconv5, econv4 & ELU \\ \hline

upconv4       & 3      & 1      & 128      & 16    & iconv5                    & ELU \\
iconv4        & 3      & 1      & 128      & 8     & $\uparrow$upconv4, econv3 & ELU \\
disp4         & 3      & 1      & 1        & 1     & iconv4                    & Sigmoid \\ \hline

upconv3       & 3      & 1      & 64       & 8     & iconv4                    & ELU \\
iconv3        & 3      & 1      & 64       & 4     & $\uparrow$upconv3, econv2 & ELU \\
disp3         & 3      & 1      & 1        & 1     & iconv3                   & Sigmoid  \\ \hline

upconv2       & 3      & 1      & 32       & 4     & iconv3                    & ELU \\
iconv2        & 3      & 1      & 32       & 2     & $\uparrow$upconv2, econv1 & ELU \\
disp2         & 3      & 1      & 1        & 1     & iconv2                    & Sigmoid \\ \hline

upconv1       & 3      & 1      & 16       & 2     & iconv2                    & ELU \\
iconv1        & 3      & 1      & 16       & 1     & $\uparrow$upconv1         & ELU \\
disp1         & 3      & 1      & 1        & 1     & iconv1                    & Sigmoid \\ \hline
\end{tabular}  \\

\begin{tabular}[t]{|l|l|l|l|l|l|l|}

\hline
\multicolumn{7}{|l|}{\textbf{Pose Decoder}} \\
\hline
\textbf{layer} & \textbf{k} & \textbf{s} & \textbf{chns} & \textbf{res} & \textbf{input} & \textbf{activation}      \\ \hline
pconv0         & 1      & 1      & 256      & 32     & econv5                    & ReLU \\ \hline
pconv1         & 3      & 1      & 256      & 32     & pconv0 & ReLU\\ \hline %
pconv2         & 3      & 1      & 256      & 32    & pconv1                     & ReLU\\
pconv3         & 1      & 1      & 6        & 32    & pconv3                     & - \\ \hline
\end{tabular} 
\end{tabular} 
 }
  \vspace{2pt}
  \caption{\textbf{Our network architecture} Here \textbf{k} is the kernel size, \textbf{s} the stride, \textbf{chns} the number of output channels for each layer, \textbf{res} is the downscaling factor for each layer relative to the input image, and \textbf{input} corresponds to the input of each layer where $\uparrow$ is a $2\times$ nearest-neighbor upsampling of the layer.}
  \vspace{40pt}
\label{tab:network}
\end{table}

\definecolor{light-gray}{gray}{0.85}
\renewcommand{\hlinegray}{\arrayrulecolor{light-gray}\hline\arrayrulecolor{black}}
\renewcommand{\clinegrayone}{\arrayrulecolor{light-gray}\cline{2-14}\arrayrulecolor{black}}

\setlength\tabcolsep{4pt} %

\begin{table*}[!t]
  \centering
  \resizebox{\textwidth}{!}
{
    \footnotesize
    \begin{tabular}{|l|l||c|c||c|c|c||c|c|c|c|c|c|c|}
      \hline
      & & 
      \begin{tabular}{@{}c@{}}Auto- \\ masking\end{tabular} & 
      \begin{tabular}{@{}c@{}}Min. \\ reproj.\end{tabular} & 
      \begin{tabular}{@{}c@{}}Full-res \\ multi-scale\end{tabular} &
      Encoder &
      Pretrained &  
      \cellcolor{col1}Abs Rel & \cellcolor{col1}Sq Rel & \cellcolor{col1}RMSE  &
      \cellcolor{col1}\begin{tabular}{@{}c@{}}RMSE \\ log\end{tabular} & 
      \cellcolor{col2}$\delta <$ 1.25 & \cellcolor{col2}$\delta < 1.25^{2}$ & \cellcolor{col2}$\delta < 1.25^{3}$ \\
    
      \hline  %
    
      (a) &  Baseline    &   &   &  & R18 & \checkmark &        
      0.140 &   1.610 &   5.512 &   0.223 &   0.852 &   0.946 &   0.973 \\  %

      \clinegrayone

      & Monodepth2 w/o min~reprojection &
      \checkmark &  & \checkmark & R18&  \checkmark & 
      0.117 &   0.878 &   4.846 &   0.196 &   0.870 &   0.957 &   0.980 \\ %
      \clinegrayone

      & Monodepth2 w/o auto-masking &  & \checkmark & \checkmark& R18  & \checkmark &
      0.120 &   1.097 &   5.074 &   0.197 &   0.872 &   0.956 &   0.979 \\  %
      \clinegrayone

      & Monodepth2 w/o full-res m.s. & \checkmark & \checkmark & & R18  &  \checkmark & 
      0.117 &   0.866 &   4.864 &   0.196 &   0.871 &   0.957 &   {\bf 0.981}  \\  %
      \clinegrayone

     & Monodepth2 w/o SSIM & \checkmark & \checkmark & \checkmark & R18  & \checkmark &   
       0.118 &   {\bf 0.853} &   {\bf 4.824} &   0.198 &   0.868 &   0.956 &   0.980 \\ %
     \clinegrayone

      & Monodepth2 with \cite{zhou2017unsupervised}'s mask &  & \checkmark & \checkmark & R18 & \checkmark &   
      0.123 &   1.177 &   5.210 &   0.200 &   0.869 &   0.955 &   0.978 \\  %
      \clinegrayone

      & \textbf{Monodepth2} (full)              &  \checkmark &  \checkmark & \checkmark &  R18 & \checkmark & 
      {\bf 0.115} &   0.903 &   4.863 &   {\bf 0.193} &   {\bf 0.877} &   {\bf 0.959} &   {\bf 0.981} \\ %

      \hline  %

      (b) & Baseline w/o pt    &  & & &  R18 &  & 
      0.150 &   1.585 &   5.671 &   0.234 &   0.827 &   0.938 &   0.971 \\ %
      \clinegrayone

      & Monodepth2 w/o pt or auto-masking    &  & \checkmark & \checkmark &R18 &  & 
      0.138 &   1.197 &   5.369 &   0.215 &   0.842 &   0.945 &   0.975 \\  %
      \clinegrayone

      & Monodepth2 w/o pt or min~reproj & \checkmark &  & \checkmark & R18 &    & 
      0.133 &   {\bf 1.021} &   5.219 &   0.214 &   0.841 &   0.945 &   0.976 \\ %
      \clinegrayone

      & Monodepth2 w/o pt or full-res m.s. & \checkmark & \checkmark &  & R18  & & 
      {\bf 0.131} &   1.030 &   5.206 &   {\bf 0.210} &   {\bf 0.846} &   {\bf 0.948} &   {\bf 0.978} \\  %
      \clinegrayone

      & \textbf{Monodepth2}  w/o pt              &  \checkmark &  \checkmark &   \checkmark & R18 & & 
      0.132 &   1.044 &   {\bf 5.142} &   {\bf 0.210} &    0.845 &   {\bf 0.948} &   0.977 \\  %

      \hline  %
    
    (c)& \textbf{Monodepth2}  ResNet18 w/o pt              &  \checkmark &  \checkmark &   \checkmark & R18 & & 
      0.132 &   1.044 &   5.142 &   0.210 &    0.845 &   0.948 &   0.977 \\  %
     \clinegrayone
      & \textbf{Monodepth2} ResNet18            &  \checkmark &  \checkmark &  \checkmark & R18  & \checkmark & 
      0.115 &   0.903 &   4.863 &  0.193 &    0.877 &   0.959 &   0.981 \\ %

      \clinegrayone
      
       &  \textbf{Monodepth2} ResNet 50 w/o pt &  \checkmark& \checkmark& \checkmark & R50 &  & 
        0.131 &   1.023 &   5.064 &   0.206 &   0.849 &   0.951 &   0.979 \\ %
       
       \clinegrayone
       & \textbf{Monodepth2} ResNet 50 & \checkmark &  \checkmark&   \checkmark & R50 &  \checkmark &  
      \textbf{0.110} &   \textbf{0.831} &   \textbf{4.642} &   \textbf{0.187} &   \textbf{0.883} &   \textbf{0.962} &   \textbf{0.982} \\ %
        \hline

      \end{tabular}
  }
    \vspace{0pt}
      \caption{
      \textbf{Ablation.} 
      Results for different variants of our model (\textbf{Monodepth2}) with monocular training (except where specified) on KITTI 2015~\cite{Geiger2012CVPR}.
      }
      \vspace{-2pt}
\label{tab:kitti_eigen_ablation_sup}
\end{table*}

\begin{figure*}
  \centering
  \includegraphics[width=0.9\textwidth]{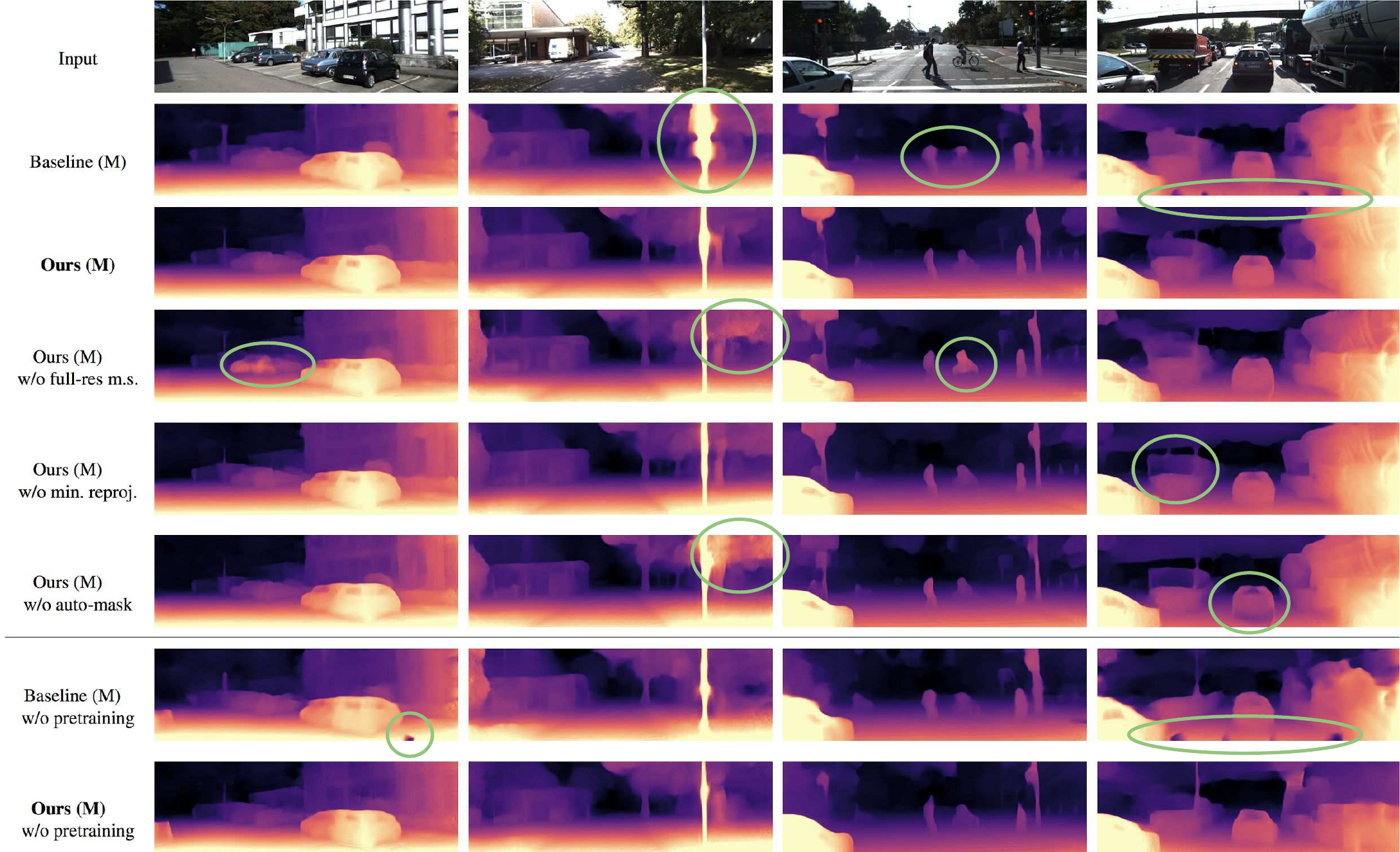} 
  \caption{\textbf{Qualitative ablation study.} We can see that our model with all components added result in the smallest amount of depth artifacts. `Baseline (M)' is our model without our full-resolution multi-scale appearance loss, minimum reprojection loss, or auto-masking loss.}
  \label{fig:kitti_eigen_qual_compar}    
\end{figure*}

\section{Network Details}
\label{sec:network_details}
Except where stated, for all experiments we use a standard ResNet18 \cite{he2016deep} encoder for both depth and pose networks.
Our pose encoder is modified to accept a pair of frames, or six channels, as input.
Our pose encoder therefore has convolutional weights in the first layer of shape $6 \times 64 \times 3 \times 3$, instead of the ResNet default of $3 \times 64 \times 3 \times 3$. 
When using pretrained weights for our pose encoder, we duplicate the first pretrained filter tensor along the channel dimension to make a filter of shape $6 \times 64 \times 3 \times 3$.
This allows for a six-channel input image.
All weights in this new expanded filter are divided by 2 to make the output of the convolution in the same numerical range as the original, one-image ResNet.
In Table \ref{tab:network} we describe the parameters of each layer used in our depth decoder and pose network.
Our pose network is larger and deeper than previous works \cite{zhou2017unsupervised, wang2017learning}, and we only feed two frames at a time to the pose network in contrast to previous works which use three \cite{zhou2017unsupervised, wang2017learning} or more for their depth estimation experiments.
In Section \ref{sec:pose_encoder_comparison} we validate the benefit of bringing additional parameters to the pose network.

\begin{table*}[t]
  \centering
  \resizebox{0.7\textwidth}{!}{
  \begin{tabular}{|l|c||c|c|c|c|c|c|c|}
  \hline
  Method & Train & \cellcolor{col1}Abs Rel & \cellcolor{col1}Sq Rel & \cellcolor{col1}RMSE  & \cellcolor{col1}RMSE log & \cellcolor{col2}$\delta < 1.25 $ & \cellcolor{col2}$\delta < 1.25^{2}$ & \cellcolor{col2}$\delta < 1.25^{3}$\\
  \hline 
Zhou \cite{zhou2017unsupervised}\textdagger & M & 0.176 & 1.532 & 6.129 & 0.244 & 0.758 & 0.921 & 0.971\\
Mahjourian \cite{mahjourian2018unsupervised} & M & 0.134 & 0.983 & 5.501 & 0.203 & 0.827 & 0.944 & 0.981\\
GeoNet \cite{geonet2018} & M  & 0.132 & 0.994 & 5.240 & 0.193 & 0.833 & 0.953 & 0.985\\
DDVO \cite{wang2017learning} & M  & 0.126 & 0.866 & 4.932 & 0.185 & 0.851 & 0.958 & 0.986\\ 
Ranjan \cite{ranjan2018adversarial}  & M & 0.123 & 0.881 & 4.834 & 0.181 & 0.860 & 0.959 & 0.985\\
EPC++ \cite{luo2018every} & M & 0.120 & 0.789 & 4.755 & 0.177 & 0.856 & 0.961 & 0.987\\
\textbf{Monodepth2} w/o pretraining & M &   \underline{0.112} &   \underline{0.715} &   \underline{4.502} &   \underline{0.167} &   \underline{0.876} &   \underline{0.967} &   \underline{0.990}  \\ %
\textbf{Monodepth2} & M &   {\bf 0.090}  &   {\bf 0.545}  &   {\bf 3.942}  &   {\bf 0.137}  &   {\bf 0.914}  &   {\bf 0.983}  &   {\bf 0.995} \\ \hline

Monodepth \cite{godard2017unsupervised} & S & 0.109 & 0.811 & 4.568 & 0.166 & 0.877 & 0.967 & 0.988\\
3net \cite{poggi20183net} (VGG) & S  &   0.119 &   0.920 &   4.824 &   0.182 &   0.856 &   0.957 &   0.985 \\ 
3net \cite{poggi20183net} (ResNet 50) & S  &   0.102 &   0.675 &   4.293 &   0.159 &   0.881 &   0.969 & \underline{0.991} \\ 
SuperDepth \cite{pillai2018superdepth} + pp & S & \underline{0.090} & \underline{0.542} & \underline{3.967} & \underline{0.144} & \underline{0.901} & \underline{0.976} & \textbf{0.993} \\

{\bf Monodepth2} w/o pretraining& S &   0.110  &   0.849  &   4.580  &   0.173  &   0.875  &   0.962  &   0.986  \\
{\bf Monodepth2}& S &   \textbf{0.085}  &   \textbf{0.537}  &   \textbf{3.868}  &   \textbf{0.139}  &   \textbf{0.912}  &   \textbf{0.979}  &   \textbf{0.993}  \\

\hline

Zhan FullNYU \cite{zhanst2018} & D*MS  & 0.130 & 1.520 & 5.184 & 0.205 & 0.859 & 0.955 & 0.981\\ 
EPC++ \cite{luo2018every} & MS & 0.123 & 0.754 & 4.453 & 0.172 & 0.863 & 0.964 & \underline{0.989}\\
{\bf Monodepth2} w/o pretraining& MS & 
 \underline{0.107} & \underline{0.720} & \underline{4.345} & \underline{0.161} & \underline{0.890} & \underline{0.971} & \underline{0.989} \\
 
{\bf Monodepth2}& MS & 
 \textbf{0.080} & \textbf{0.466} & \textbf{3.681} & \textbf{0.127} & \textbf{0.926} & \textbf{0.985} & \textbf{0.995} \\ 
 
 \hline

  \end{tabular}
  }\hfill
  \vspace{4pt}
    \raisebox{3pt}{
  \begin{minipage}[c]{0.28\textwidth}
  \caption{\textbf{KITTI improved ground truth.} Comparison to existing methods on KITTI 2015 \cite{Geiger2012CVPR} using 93\% of the Eigen split and the improved ground truth from \cite{uhrig2017sparse}. Baseline methods were evaluated using their provided disparity files, which were either available publicly or from private communication with the authors.
  \vspace{8pt}  \newline 
  {\footnotesize
    \textbf{Legend} \newline 
  D* -- Auxiliary depth supervision \newline 
  S -- Self-supervised stereo supervision \newline 
  M -- Self-supervised mono supervision \newline
  \textdagger~ -- Newer results from the respective online implementations. \newline
  + pp -- With post-processing}
  \label{tab:kitti_improved_gt}}
  \end{minipage}}

\end{table*}

\section{Additional Ablation Experiments}
\label{sec:additional_ablation}
In Table \ref{tab:kitti_eigen_ablation_sup} we show a full ablation study on our algorithm, turning on and off different components of the system.
We confirm the finding of the main paper, that all our components together gives the highest quality model, and that pretraining helps.
We observe in Table \ref{tab:kitti_eigen_ablation_sup} (d) that our results with ResNet 50 are even higher than our ResNet18 models.
ResNet 50 is a standard encoder used by previous works \eg \cite{godard2017unsupervised, poggi20183net}.
However, training with a 50-layer ResNet comes at the expense of longer training and test times.
In Fig. \ref{fig:kitti_eigen_qual_compar} we show additional qualitative results for the monocular trained variants of our model from Table \ref{tab:kitti_eigen_ablation_sup}.
We observe `depth holes' in both non-pretrained and pretrained versions of the baseline model compared to ours.

\section{Additional Evaluation}

\subsection{Improved Ground Truth}
As mentioned in the main paper, the evaluation method introduced by Eigen \cite{eigen2015predicting} for KITTI uses the reprojected LIDAR points but does not handle occlusions, moving objects, or the fact that the car is moving. \cite{uhrig2017sparse} introduced a set of high quality depth maps for the KITTI dataset, making use of 5 consecutive frames and handling moving objects using the stereo pair. 
This improved ground truth depth is provided for 652 (or 93\%) of the 697 test frames contained in the Eigen test split \cite{eigen2015predicting}. 
We evaluate our results on these 652 improved ground truth frames and compare to existing published methods without having to retrain each method, see Table~\ref{tab:kitti_improved_gt}.
We present results for all other methods for which we have obtained predictions from the authors.
We use the same error metrics from the standard evaluation, and clip the predicted depths to 80 meters to match the Eigen evaluation. 
We evaluate on the full image and do not crop, unlike with the Eigen evaluation. 
We can see that our method still significantly outperforms all previously published methods on all metrics. 
While Superdepth \cite{pillai2018superdepth} comes a close second to our algorithm in the S category, they are run at high resolution ($1024\times384$ vs.~our $640\times192$), and in Table~\ref{tab:kitti_eigen} we show that at higher resolutions our model's performance also increases.

\begin{table*}
  \centering
  \resizebox{0.7\textwidth}{!}{
  \begin{tabular}{|l||c||c|c|c|c|c|c|c|}
  \hline
  Method & \cellcolor{col1}$\sigma_{scale}$ & \cellcolor{col1}Abs Rel & \cellcolor{col1}Sq Rel & \cellcolor{col1}RMSE  & \cellcolor{col1}RMSE log & \cellcolor{col2}$\delta < 1.25 $ & \cellcolor{col2}$\delta < 1.25^{2}$ & \cellcolor{col2}$\delta < 1.25^{3}$\\
  \hline 
Zhou \cite{zhou2017unsupervised}\textdagger & 0.210  & 0.258 & 2.338 & 7.040 & 0.309 & 0.601 & 0.853 & 0.940\\
Mahjourian \cite{mahjourian2018unsupervised} & 0.189  & 0.221 & 1.663 & 6.220 & 0.265 & 0.665 & 0.892 & 0.962\\
GeoNet \cite{geonet2018} & 0.172  & 0.202 & 1.521 & 5.829 & 0.244 & 0.707 & 0.913 & 0.970\\
Ranjan \cite{ranjan2018adversarial} & 0.162 & 0.188 & 1.298 & 5.467 & 0.232 & 0.724 & 0.927 & 0.974\\
EPC++ \cite{luo2018every} & 0.123 & 0.153 & 0.998 & 5.080 & 0.204 & 0.805 & 0.945 & 0.982\\
DDVO \cite{wang2017learning} & 0.108  &  0.147 & 1.014 & 5.183 & 0.204 & 0.808 & 0.946 & 0.983 \\ 
 \textbf{Monodepth2} & \textbf{0.093} &   \textbf{0.109} &  \textbf{0.623} &   \textbf{4.136} &   \textbf{0.154} &   \textbf{0.873} &   \textbf{0.977} &   \textbf{0.994} \\ 
 \hline
  \end{tabular}
  }
  \vspace{4pt}
  \raisebox{3pt}{
  \begin{minipage}[c]{0.28\textwidth}
  \caption{\textbf{Single scale monocular evaluation.} Comparison to existing monocular supervised methods on KITTI 2015 \cite{Geiger2012CVPR} using the Eigen split with improved ground truth from \cite{uhrig2017sparse} using a \emph{single} scale for each method.
  \textdagger~~indicates newer results from the online implementation.
  }
  \label{tab:kitti_eigen_single_scale}
  \end{minipage}}
  
\end{table*}

\begin{table}
\centering
\resizebox{1.0\columnwidth}{!}{
\begin{tabular}{| c | c | c | c | c | c | c}
\hline
Method & Train & \cellcolor{col1} SILog & \cellcolor{col1} sqErrorRel & \cellcolor{col1} absErrorRel  & \cellcolor{col1} iRMSE \\ \hline
DORN \cite{fu2018deep} & D & 11.77 & 2.23  & 8.78  & 12.98 \\
DABC \cite{li2018deep}& D & 14.49 & 4.08  & 12.72  & 15.53 \\
APMoE \cite{kong2018pixel} & D & 14.74 & 3.88  & 11.74  & 15.63 \\
CSWS \cite{li2018monocular}& D & 14.85 & 3.48  & 11.84  & 16.38 \\
DHGRL \cite{zhang2018deep}& D & 15.47 & 4.04  & 12.52  & 15.72 \\ \hline
Monodepth \cite{godard2017unsupervised} & S & 22.02	& 20.58 & 	17.79 &	21.84\\
{\bf Monodepth2} & M & 15.57 & 4.52 & 12.98 & 16.70 \\
{\bf Monodepth2} & MS & 15.07 & 4.16 & 11.64 & 15.27 \\
{\bf Monodepth2} (ResNet 50) & MS  & 14.41 & 3.67 & 11.22 & 14.73\\\hline
\end{tabular}
}
  \vspace{4pt}
  \caption{\textbf{KITTI depth prediction benchmark.}  Comparison of our monocular plus stereo approaches to fully supervised methods on the KITTI depth prediction benchmark \cite{kittidepthserver}. D indicates models that were trained with ground truth depth supervision, while M and S are monocular and stereo self-supervision respectively. 
  }
\label{tab:kitti_eval_server}
\end{table}

\subsection{Single-Scale Evaluation}
\label{sec:single_scale_eval}

Our monocular trained approach, like all self-supervised baselines, has no guarantee of producing results with a metric scale. 
Nonetheless, we anticipate that there could be value in estimating depth-outputs that are, without special measures, consistent with each other across all predictions. 
In \cite{zhou2017unsupervised}, the authors independently scale each predicted depth map by the ratio of the median of the ground truth and predicted depth map -- for \emph{each} individual test image. 
This is in contrast to stereo based training where the scale is known and as a result no additional scaling is required during the evaluation \eg \cite{garg2016unsupervised,godard2017unsupervised}. 
This per-image depth scaling hides unstable scale estimation in both depth and pose estimation and presents a best-case scenario for the monocular training case. 
If a method outputs wildly varying scales for each sequence, then this evaluation protocol will hide the issue. 
This gives an unfair advantage over stereo trained methods that do not perform per-image scaling.

We thus modified the original protocol to instead use a single scale for all predicted depth maps of each method. For each method, we compute this single scale by taking the median of all the individual ratios of the depth medians on the \emph{test} set. 
While this is still not ideal as it makes use of the ground truth depth, we believe it to be fairer and representative of the performance of each method. 
We also calculated the standard deviation $\sigma_{scale}$ of the individual scales, where lower values indicate more consistent output depth map scales. 
As can be seen in Table~\ref{tab:kitti_eigen_single_scale}, our method outperforms previously published self-supervised monocular methods, especially in the near range depth values \ie $\delta < 1.25$, and is more stable overall.

\subsection{KITTI Evaluation Server Benchmark}
\label{sec:kitt_online_bench}
Here, we report the performance of our self-supervised monocular plus stereo model on the online KITTI single image depth prediction benchmark evaluation server \cite{kittidepthserver}.
\cite{kittidepthserver} uses a different split of the data, which is not the same as the Eigen split.
As a result, we train a new model on the provided training data. 
At the time of writing, there were no published self-supervised approaches among the submissions on the leaderboard.
Despite not using any ground truth data during training, our monocular only predictions are competitive with fully supervised methods, see Table \ref{tab:kitti_eval_server}.
Adding stereo data and a more powerful encoder at training time results in even better performance ({\bf Monodepth2} (ResNet50)). 

Because the evaluation server does not do median scaling (required for monocular methods), we needed a way to find the correct scaling for our mono-only model, which makes unscaled predictions. 
We make predictions with our mono-model on 1,000 images from the KITTI training set which have ground truth depths available, and for each of the 1,000 images we find the scale factor which best align the depth maps \cite{zhou2017unsupervised}.
Finally, we take the median of these 1,000 scale factors as the single factor which we use to scale all predictions from our mono model.
Note that, to remain true to our `self-supervised' philosophy, we never do any other form of validation, model selection or parameter tuning using ground truth depths.
For comparison, we trained a version of the original Monodepth \cite{godard2017unsupervised} using the online code\footnote{\url{https://github.com/mrharicot/monodepth}} on the same benchmark split.

\section{Additional Qualitative Comparisons}
\label{sec:additional_results}
\begin{figure}
  \centering
  \resizebox{1.0\columnwidth}{!}{
  \newcommand{\turnheightnew}{0.195\columnwidth}

\centering

\begin{tabular}{@{\hskip 1mm}c@{\hskip 1mm}c@{\hskip 1mm}c@{\hskip 1mm}c@{\hskip 1mm}c@{}}

\Large{Input} & \Large{Zhou~\ea\cite{zhou2017unsupervised}}  & \Large{DDVO~\cite{wang2017learning}}  & \Large{MD2 M} & \Large{Ground truth} \\

\includegraphics[height=\turnheightnew]{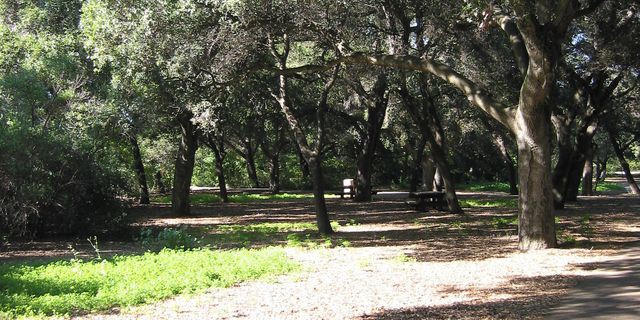} &
\includegraphics[height=\turnheightnew]{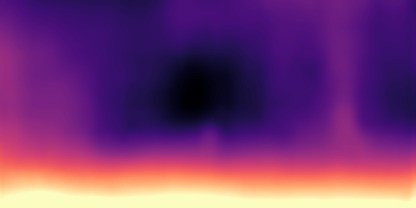} &
\includegraphics[height=\turnheightnew]{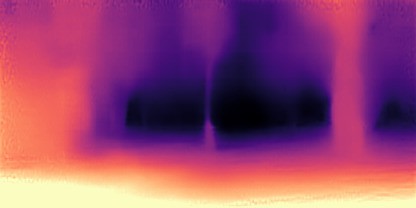} &
\includegraphics[height=\turnheightnew]{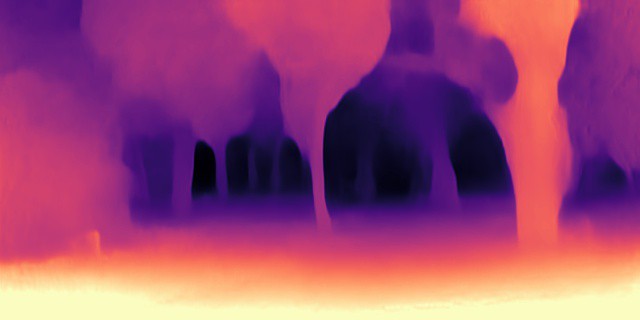} &
\includegraphics[height=\turnheightnew]{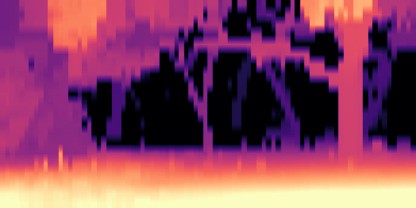}\\

\includegraphics[height=\turnheightnew]{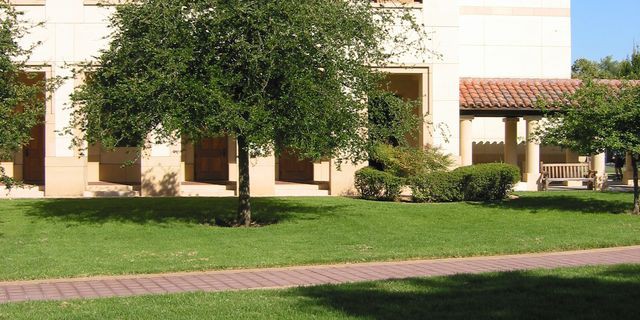} &
\includegraphics[height=\turnheightnew]{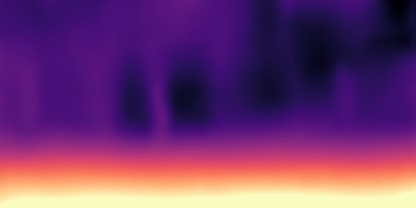} &
\includegraphics[height=\turnheightnew]{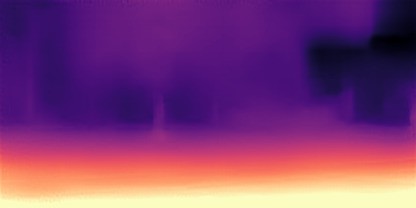} &
\includegraphics[height=\turnheightnew]{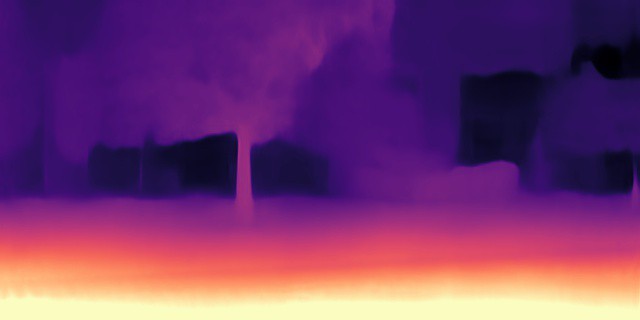} &
\includegraphics[height=\turnheightnew]{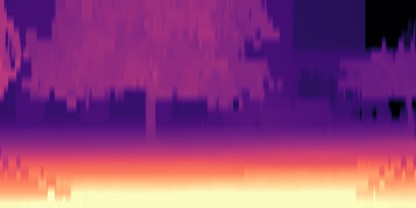}\\

\includegraphics[height=\turnheightnew]{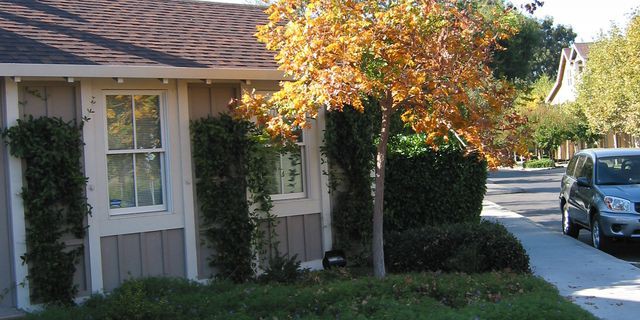} &
\includegraphics[height=\turnheightnew]{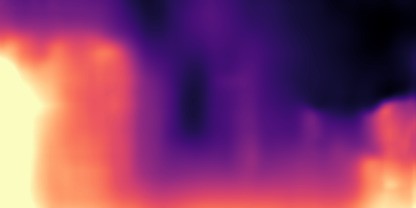} &
\includegraphics[height=\turnheightnew]{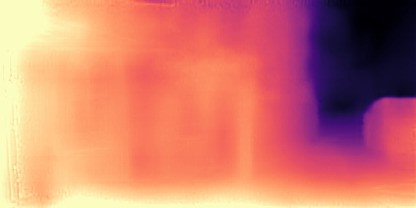} &
\includegraphics[height=\turnheightnew]{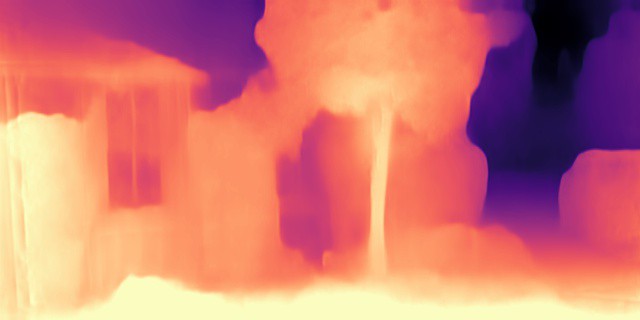} &
\includegraphics[height=\turnheightnew]{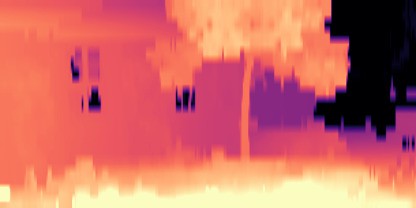}\\

\includegraphics[height=\turnheightnew]{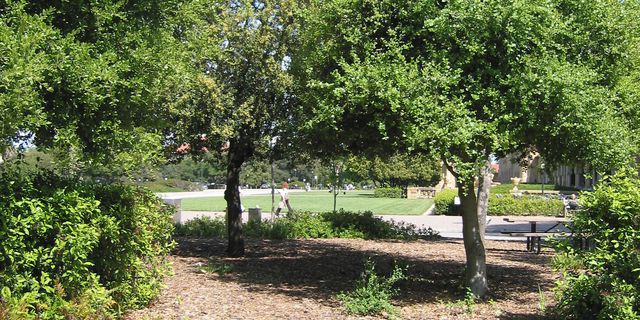} &
\includegraphics[height=\turnheightnew]{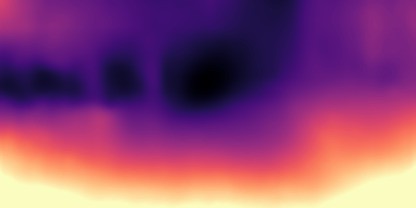} &
\includegraphics[height=\turnheightnew]{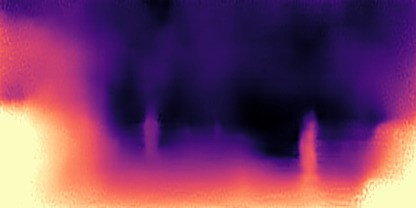} &
\includegraphics[height=\turnheightnew]{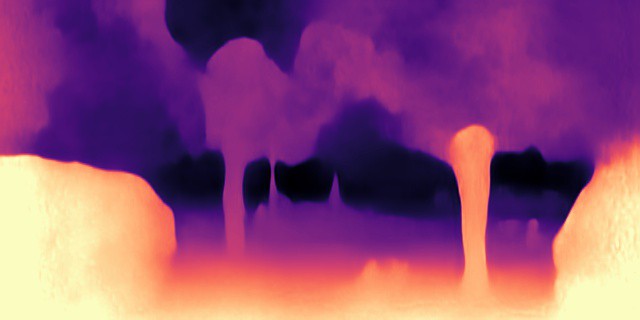} &
\includegraphics[height=\turnheightnew]{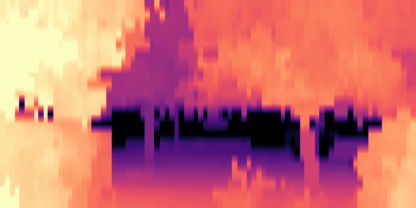}\\

\includegraphics[height=\turnheightnew]{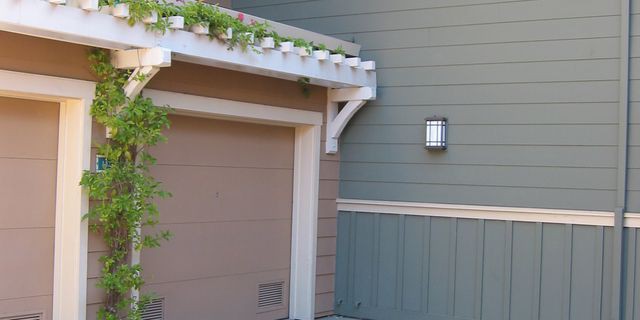} &
\includegraphics[height=\turnheightnew]{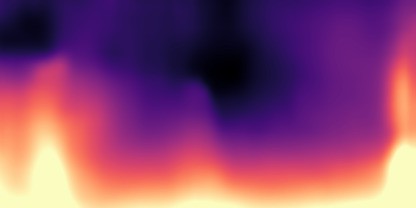} &
\includegraphics[height=\turnheightnew]{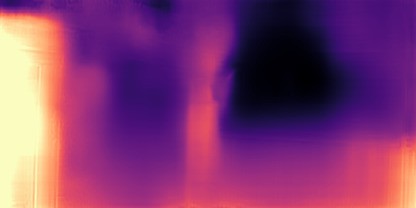} &
\includegraphics[height=\turnheightnew]{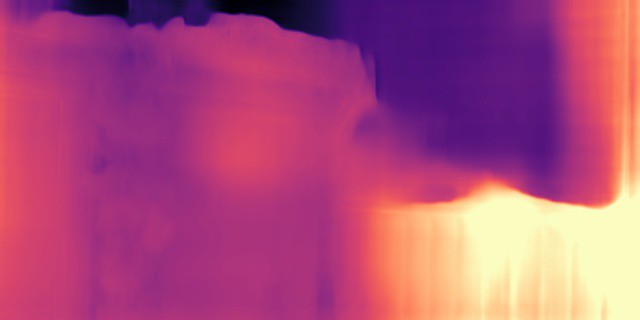} &
\includegraphics[height=\turnheightnew]{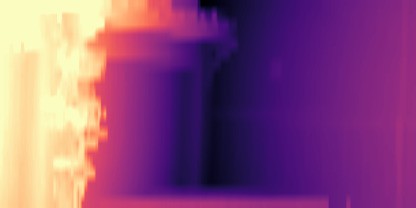}\\

\end{tabular}
 }
  \caption{\textbf{Additional Make3D results.} Our model (MD2 M) trained on KITTI results in plausible depths, predicting more detail than existing monocular methods. The last row is an interesting failure for all methods as it contains an image that is very different than those from the KITTI training set.}
  \label{tab:make3d_sup}    
\end{figure}

\begin{figure*}
  \centering
  \resizebox{0.9\textwidth}{!}{
  \newcommand{\turnheightnew}{0.25\columnwidth}

\centering

\begin{tabular}{@{\hskip 2mm}c@{\hskip 2mm}c@{\hskip 2mm}c@{\hskip 2mm}c@{\hskip 2mm}c@{}}

{\rotatebox{90}{\hspace{6mm}Input}} &
\includegraphics[height=\turnheightnew]{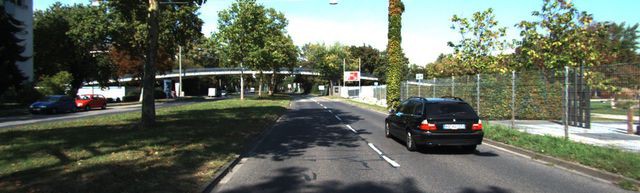} &
\includegraphics[height=\turnheightnew]{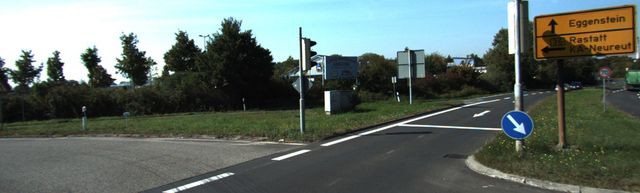} &
\includegraphics[height=\turnheightnew]{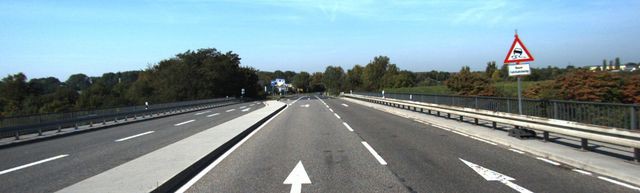} &
\includegraphics[height=\turnheightnew]{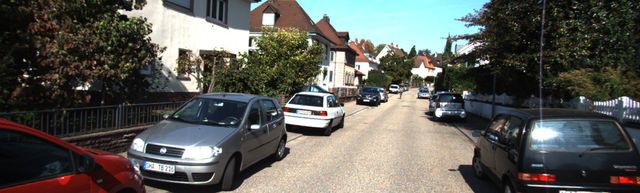}\\

{\rotatebox{90}{\hspace{2mm}Garg~\ea~\cite{garg2016unsupervised}}} &
\includegraphics[height=\turnheightnew]{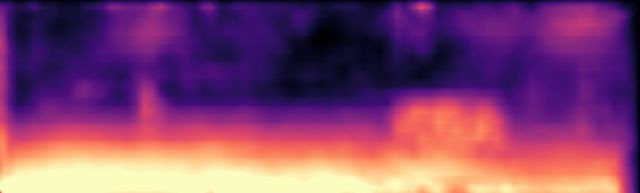} &
\includegraphics[height=\turnheightnew]{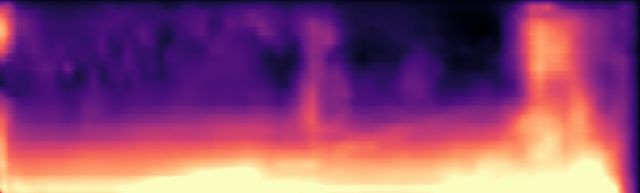} &
\includegraphics[height=\turnheightnew]{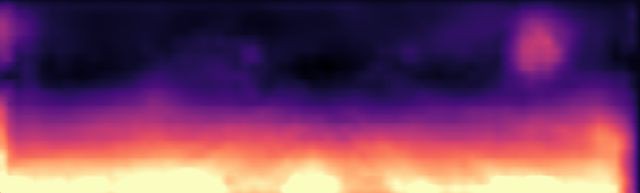} &
\includegraphics[height=\turnheightnew]{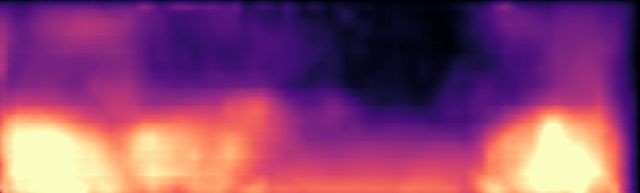} \\

{\rotatebox{90}{\hspace{1mm}Monodepth \cite{godard2017unsupervised}}} &
\includegraphics[height=\turnheightnew]{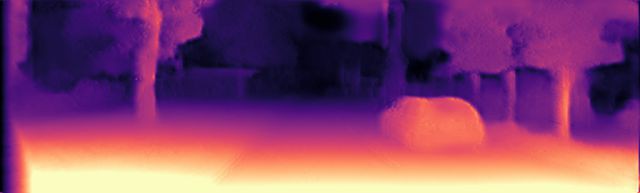} &
\includegraphics[height=\turnheightnew]{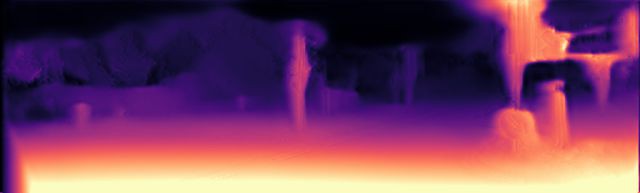} &
\includegraphics[height=\turnheightnew]{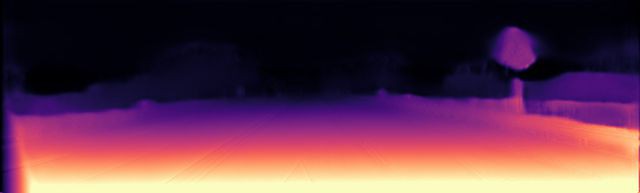} &
\includegraphics[height=\turnheightnew]{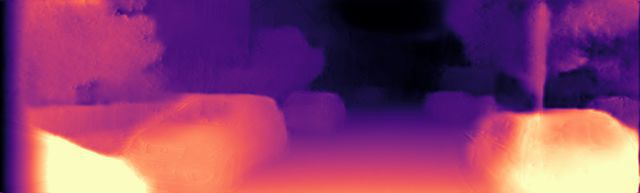} \\

{\rotatebox{90}{\hspace{2mm}Zhou \ea~\cite{zhou2017unsupervised}}} &
\includegraphics[height=\turnheightnew]{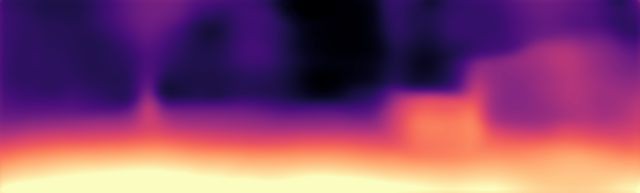} &
\includegraphics[height=\turnheightnew]{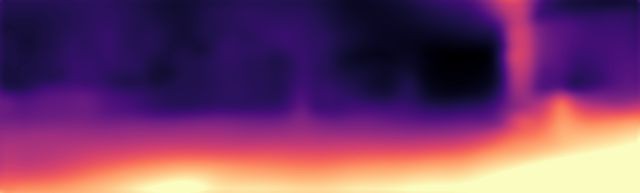} &
\includegraphics[height=\turnheightnew]{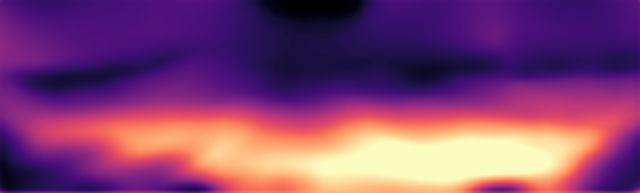} &
\includegraphics[height=\turnheightnew]{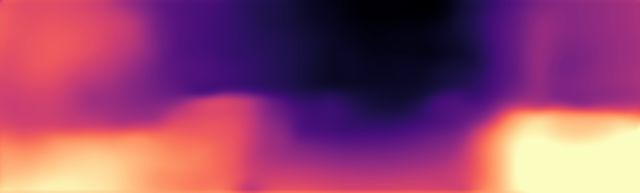} \\

{\rotatebox{90}{\hspace{2mm}Zhan \ea~\cite{zhanst2018}}} &
\includegraphics[height=\turnheightnew]{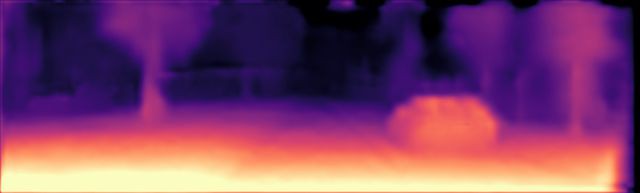} &
\includegraphics[height=\turnheightnew]{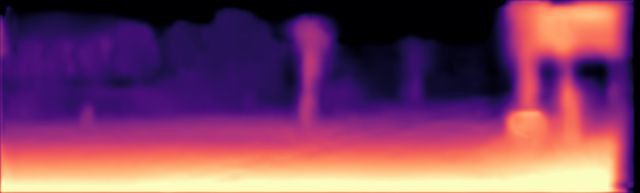} &
\includegraphics[height=\turnheightnew]{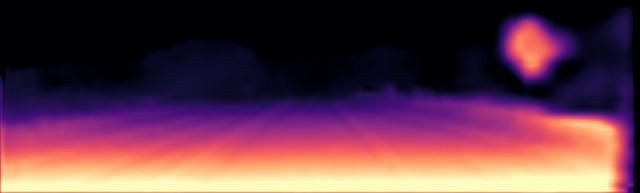} &
\includegraphics[height=\turnheightnew]{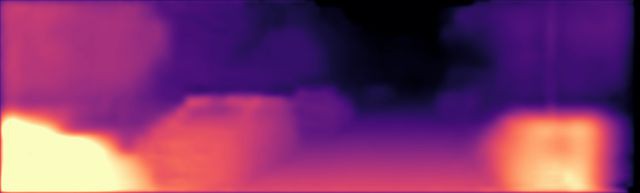} \\

{\rotatebox{90}{\hspace{4mm}DDVO~\cite{wang2017learning}}} &
\includegraphics[height=\turnheightnew]{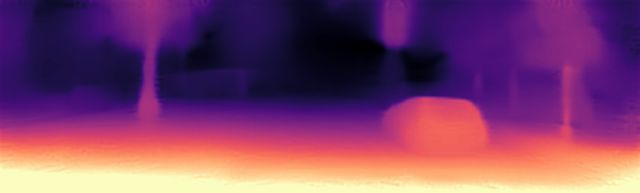} &
\includegraphics[height=\turnheightnew]{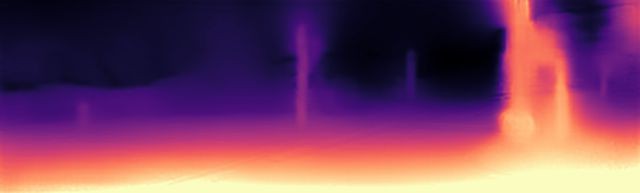} &
\includegraphics[height=\turnheightnew]{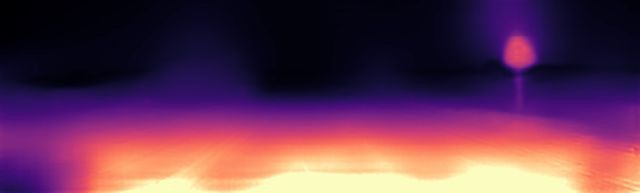} &
\includegraphics[height=\turnheightnew]{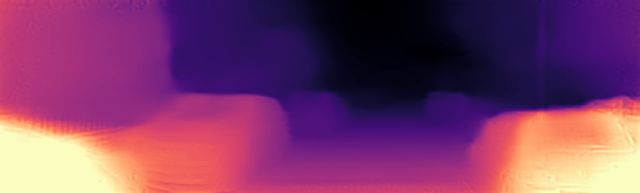} \\

{\rotatebox{90}{\hspace{0mm}\small{Mahjourian \ea~\cite{mahjourian2018unsupervised}}}} &
\includegraphics[height=\turnheightnew]{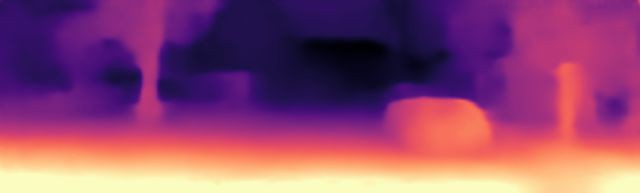} &
\includegraphics[height=\turnheightnew]{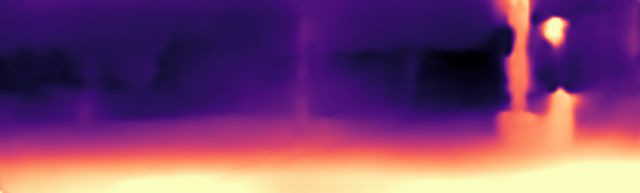} &
\includegraphics[height=\turnheightnew]{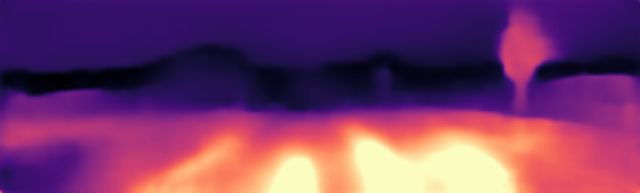} &
\includegraphics[height=\turnheightnew]{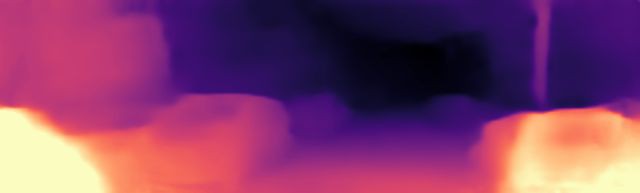} \\

{\rotatebox{90}{\hspace{4mm}GeoNet~\cite{geonet2018}}} &
\includegraphics[height=\turnheightnew]{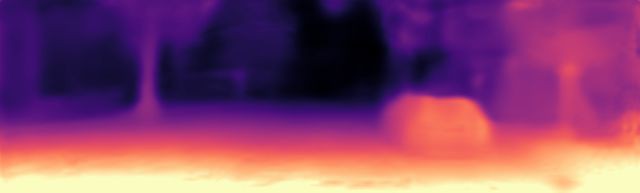} &
\includegraphics[height=\turnheightnew]{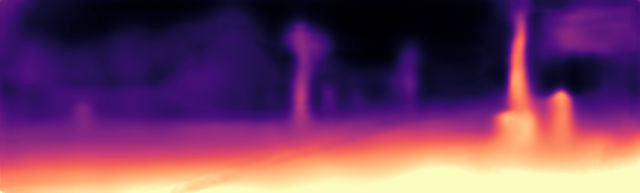} &
\includegraphics[height=\turnheightnew]{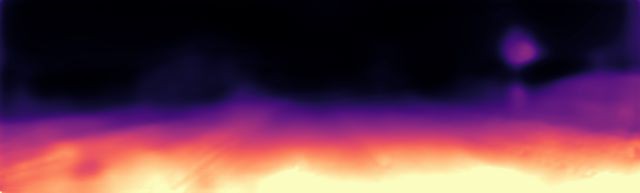} &
\includegraphics[height=\turnheightnew]{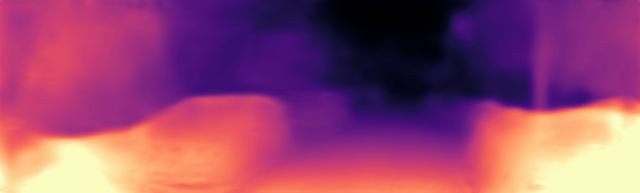} \\

{\rotatebox{90}{\hspace{4mm}Ranjan \ea~\cite{ranjan2018adversarial}}} &
\includegraphics[height=\turnheightnew]{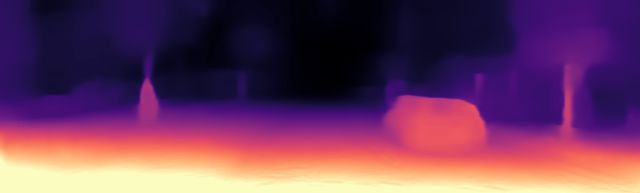} &
\includegraphics[height=\turnheightnew]{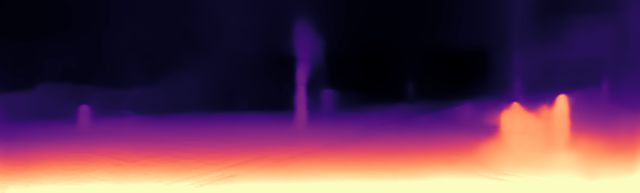} &
\includegraphics[height=\turnheightnew]{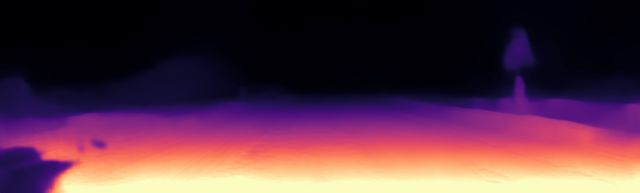} &
\includegraphics[height=\turnheightnew]{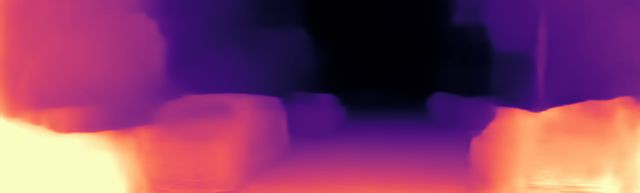} \\

{\rotatebox{90}{\hspace{4mm}EPC++~\cite{luo2018every}}} &
\includegraphics[height=\turnheightnew]{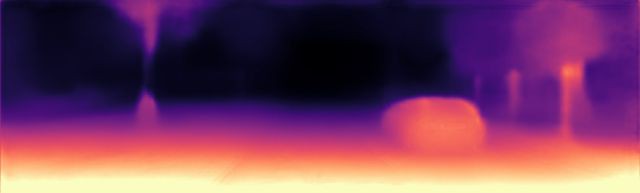} &
\includegraphics[height=\turnheightnew]{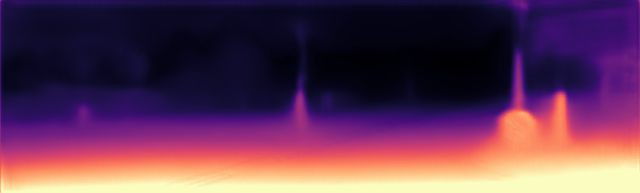} &
\includegraphics[height=\turnheightnew]{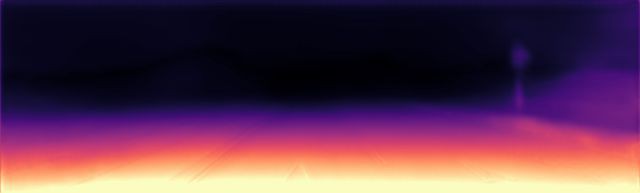} &
\includegraphics[height=\turnheightnew]{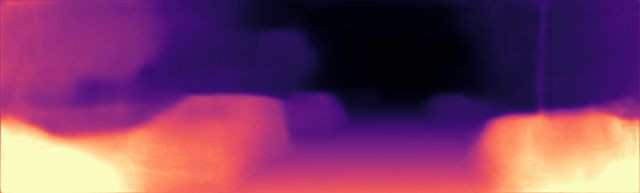} \\

{\rotatebox{90}{\hspace{5mm}3Net \cite{poggi20183net}}} &
\includegraphics[height=\turnheightnew]{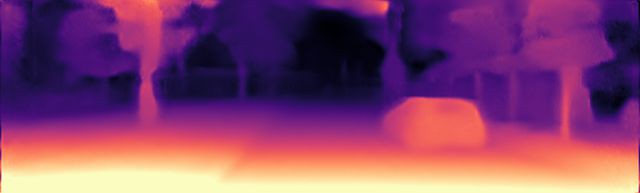} &
\includegraphics[height=\turnheightnew]{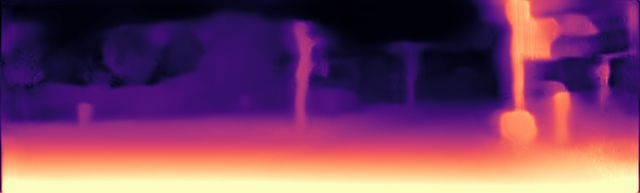} &
\includegraphics[height=\turnheightnew]{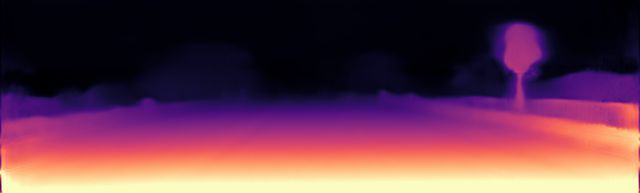} &
\includegraphics[height=\turnheightnew]{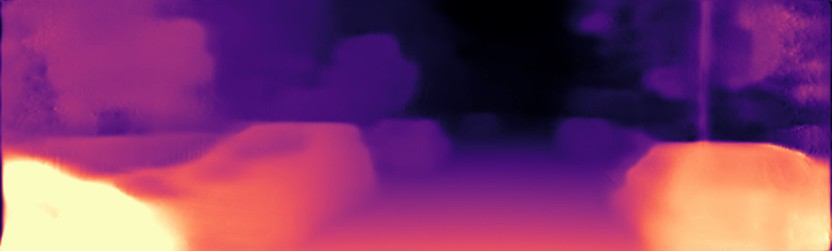} \\

{\rotatebox{90}{\hspace{5mm}Baseline M}} &
\includegraphics[height=\turnheightnew]{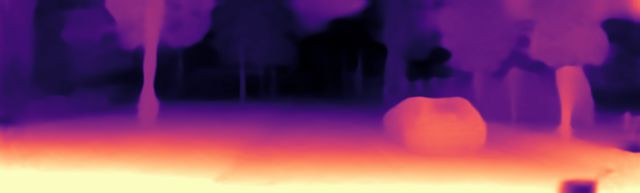} &
\includegraphics[height=\turnheightnew]{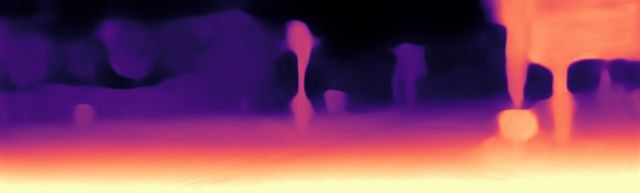} &
\includegraphics[height=\turnheightnew]{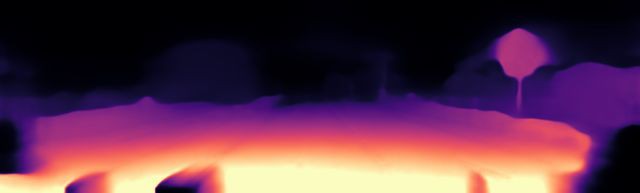} &
\includegraphics[height=\turnheightnew]{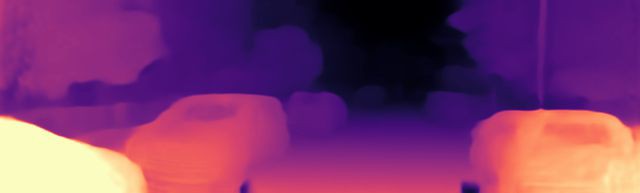} \\

{\rotatebox{90}{\hspace{5mm}\textbf{Ours M}}} &
\includegraphics[height=\turnheightnew]{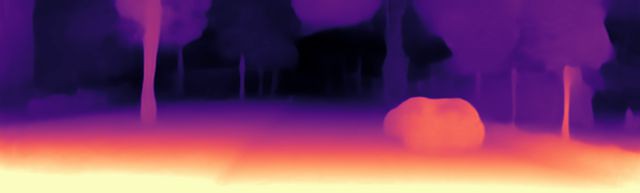} &
\includegraphics[height=\turnheightnew]{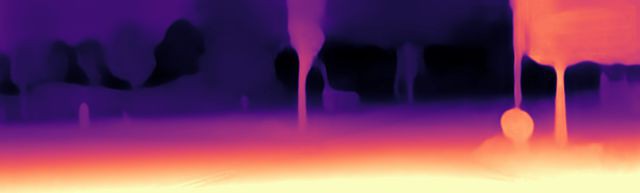} &
\includegraphics[height=\turnheightnew]{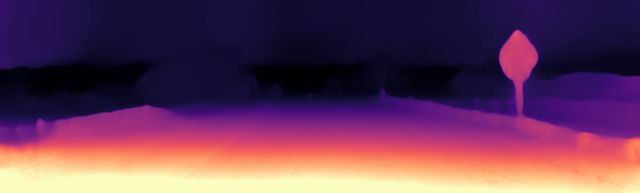} &
\includegraphics[height=\turnheightnew]{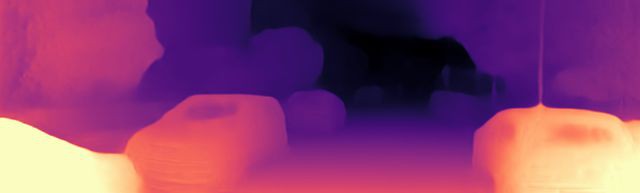} \\

{\rotatebox{90}{\hspace{5mm}\textbf{Ours S}}} &
\includegraphics[height=\turnheightnew]{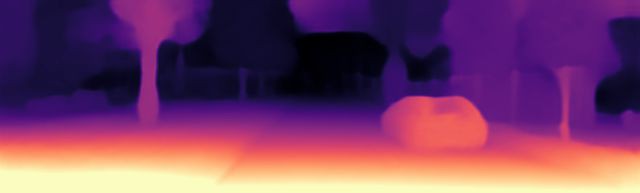} &
\includegraphics[height=\turnheightnew]{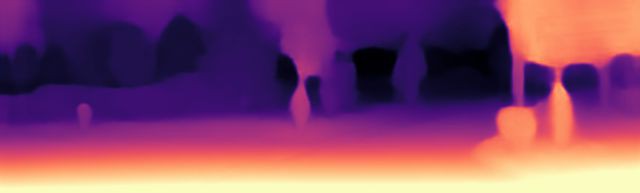} &
\includegraphics[height=\turnheightnew]{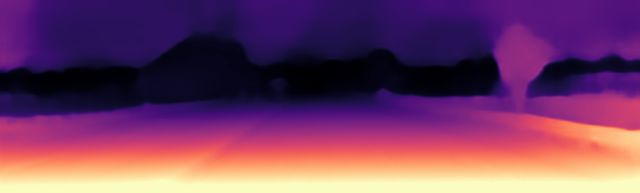} &
\includegraphics[height=\turnheightnew]{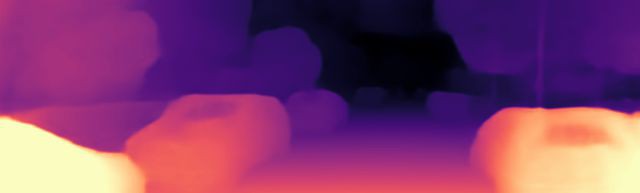} \\

{\rotatebox{90}{\hspace{4mm}\textbf{Ours MS}}} &
\includegraphics[height=\turnheightnew]{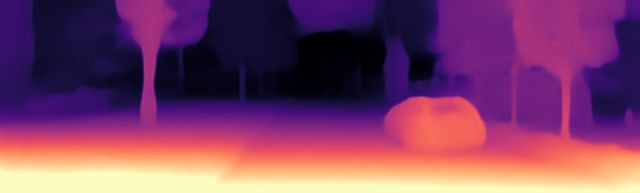} &
\includegraphics[height=\turnheightnew]{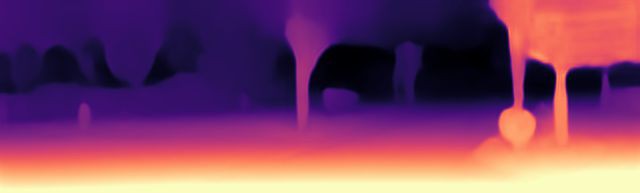} &
\includegraphics[height=\turnheightnew]{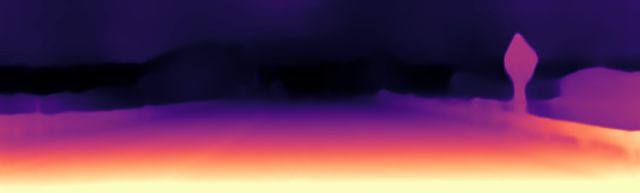} &
\includegraphics[height=\turnheightnew]{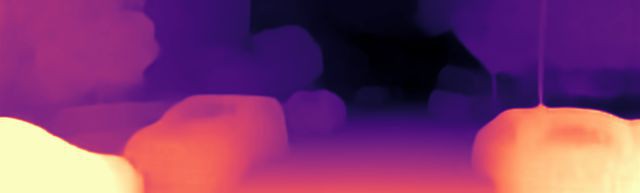} \\

\end{tabular} }
  \caption{\textbf{Additional KITTI Eigen split test results.} We can see that our approaches in the last three rows produce the sharpest depth maps. `Baseline M' is our model without our contributions.}
  \label{fig:kitti_eigen_qual_sup}    
\end{figure*}

We include additional qualitative results from the KITTI test set in Fig.~\ref{fig:kitti_eigen_qual_sup}.
We can see that our models generate higher quality outputs and do not produce `holes' in the depth maps or border artifacts that can be seen in many existing baselines \eg. \cite{zhou2017unsupervised,ranjan2018adversarial,godard2017unsupervised,zhanst2018}. 
We also show additional results from Make3D in Fig. \ref{tab:make3d_sup}.

\begin{table*}

  \centering
  \footnotesize
  \begin{tabular}{|l|c||c|c|c|c|c|c|c|}
  \hline
  Method & Train &  
  \cellcolor{col1}Abs Rel & \cellcolor{col1}Sq Rel & \cellcolor{col1}RMSE  & \cellcolor{col1}RMSE log & \cellcolor{col2}$\delta < 1.25 $ & \cellcolor{col2}$\delta < 1.25^{2}$ & \cellcolor{col2}$\delta < 1.25^{3}$\\
  \hline

    \arrayrulecolor{black}\hline

    {\bf Monodepth2} w/o pretraining & M &
    0.132 &   1.044 &   5.142 &   0.210 &   0.845 & 0.948 &   0.977 \\

    {\bf Monodepth2} w/o pretraining + pp & M &
    0.129 &   1.003 &   5.072 &   0.207 &   0.848 &   0.949 &   0.978 \\ %
    \arrayrulecolor{gray}\hline

    \textbf{Monodepth2} & M &
    0.115 &   0.903 &   4.863 &   0.193 &  0.877 &   0.959 &   0.981 \\ %

    \textbf{Monodepth2} + pp & M &
    \textbf{0.112} &   0.851 &   4.754 &   0.190 &   0.881 &   0.960 &   0.981 \\ %

    \arrayrulecolor{gray}\hline

    \textbf{Monodepth2} (1024 $\times$ 320)  &  M &
    0.115 &   0.882 &   4.701 &  0.190 &  0.879 &  0.961 &  \textbf{0.982} \\  %

    \textbf{Monodepth2} (1024 $\times$ 320)  + pp &  M &
    \textbf{0.112} &   \textbf{0.838} &   \textbf{4.607} &   \textbf{0.187} &   \textbf{0.883} &   \textbf{0.962} &   \textbf{0.982} \\ %

    \arrayrulecolor{black}\hline\hline

    {\bf Monodepth2} w/o pretraining & S  &
     0.130  &   1.144  &   5.485  &   0.232  &   0.831  &   0.932  &   0.968  \\

    {\bf Monodepth2} w/o pretraining + pp & S  &
      0.128  &   1.089  &   5.385  &   0.229  &   0.832  &   0.934  &   0.969  \\

    \arrayrulecolor{gray}\hline

    {\bf Monodepth2}& S &
      0.109 &   0.873 &   4.960 &   0.209 &   0.864 &   0.948 &   0.975 \\

    {\bf Monodepth2} + pp & S &
      0.108  &   0.842  &   4.891  &   0.207  &   0.866  &   0.949  &   0.976  \\

    \arrayrulecolor{gray}\hline

    {\bf Monodepth2}  (1024 $\times$ 320) & S &
        0.107 &   0.849 &   4.764 &   0.201 &   0.874 &   0.953 &   \textbf{0.977} \\

    {\bf Monodepth2}  (1024 $\times$ 320) + pp & S &
    \textbf{0.105}  &   \textbf{0.822}  &   \textbf{4.692}  &   \textbf{0.199}  &   \textbf{0.876}  &   \textbf{0.954}  &   \textbf{0.977}  \\

    \arrayrulecolor{black}\hline\hline

    \textbf{Monodepth2} w/o pretraining & MS &
    0.127 &   1.031 &   5.266 &   0.221 &   0.836 &   0.943 &   0.974 \\

    \textbf{Monodepth2} w/o pretraining + pp & MS &
    0.125 &   1.000 &   5.205 &   0.218 &   0.837 &   0.944 &   0.974 \\ %

    \arrayrulecolor{gray}\hline

    \textbf{Monodepth2}& MS &
    0.106 &   0.818 &   4.750 &   0.196 &   0.874 &   0.957 &   0.979 \\

    \textbf{Monodepth2} + pp & MS &
    \textbf{0.104} &   0.786 &   4.687 &   0.194 &   0.876 &   0.958 &   0.980 \\ %

    \arrayrulecolor{gray}\hline

    \textbf{Monodepth2} (1024 $\times$ 320) & MS   &
    0.106 &   0.806 &   4.630 &   0.193 &   0.876 &   0.958 &   0.980  \\ %

    \textbf{Monodepth2} (1024 $\times$ 320) + pp & MS   &
    \textbf{0.104} &   \textbf{0.775} &   \textbf{4.562} &   \textbf{0.191} &   \textbf{0.878} &   \textbf{0.959} &   \textbf{0.981} \\ %

    \arrayrulecolor{black}\hline

  \end{tabular}
  \vspace{4pt}
  \caption{\textbf{Effect of post-processing.} We observe that post-processing, originally motivated only for stereo training, also brings consistent benefits to all our monocular-trained models. Interestingly, for some metrics post-processing results in a larger quantitative gain than models trained at higher resolution. \label{tab:kitti_eigen_pp}}
\end{table*}

\begin{table*}
    
    \newcommand{\lr}{$416 \times 128$}
    \newcommand{\mr}{$640 \times 192$}
    \newcommand{\hr}{$1024 \times 320$}

  \centering
  
     \resizebox{0.9\textwidth}{!}{
  \begin{tabular}{|l||c|c|c||c|c|c|c|c|c|c||c|}
  \hline
   & Train & Resolution & Full-res multi-scale & \cellcolor{col1}Abs Rel & \cellcolor{col1}Sq Rel & \cellcolor{col1}RMSE  & \cellcolor{col1}RMSE log & \cellcolor{col2}$\delta < 1.25 $ & \cellcolor{col2}$\delta < 1.25^{2}$ & \cellcolor{col2}$\delta < 1.25^{3}$ & Train.~time (h) \\
  \hline 
  
  Monodepth2& M & \lr & \checkmark &
   0.128 &   1.087 &   5.171 &   0.204 &   0.855 &   0.953 &   0.978 & 9 \\
  
  Monodepth2& M & \mr &  \checkmark &
  {\bf 0.115} &   0.903 &   4.863 &   0.193 &   0.877 &   0.959 &   0.981 & 12 \\
  
  Monodepth2& M & \hr &    \checkmark &
  {\bf 0.115} &   {\bf 0.882} &   {\bf 4.701} &   {\bf 0.190} &   {\bf 0.879} &   {\bf 0.961} &   {\bf 0.982} & 6 + 9 \textsuperscript{\textdagger}\\  %
  
  \hline 
  
  Monodepth2& S & \lr &  \checkmark & 
     0.118  &   0.971  &   5.231  &   0.218  &   0.848  &   0.943  &   0.973  & 6 \\ 

  Monodepth2 & S & \mr &  \checkmark &
   0.109 &   0.873 &   4.960 &   0.209 &   0.864 &   0.948 &   0.975 & 8 \\  
  
  Monodepth2 & S & \hr &  \checkmark &
   \textbf{0.105}  &   \textbf{0.822}  &   \textbf{4.692}  &   \textbf{0.199}  &   \textbf{0.876}  &   \textbf{0.954}  &   \textbf{0.977} &  4 + 8 \textsuperscript{\textdagger} \\
  \hline
  
  Monodepth2& MS & \lr &  \checkmark &
    0.118 &   0.935 &   5.119 &   0.210 &   0.852 &   0.949 &   0.976 & 11 \\ %
  Monodepth2 & MS & \mr &  \checkmark &
   {\bf 0.106} &   0.818&   4.750&   0.196&   {\bf 0.874} &   0.957&   0.979& 15 \\
  
  Monodepth2& MS & \hr &  \checkmark &
    {\bf 0.106} &   {\bf 0.806} &   {\bf 4.630} &   {\bf 0.193} &   {\bf 0.876}&   {\bf 0.958} &   {\bf 0.980}&  7.5 + 10 \textsuperscript{\textdagger} \\ \hline
  
  \end{tabular}}
    \vspace{2pt}
  \caption{\textbf{Ablation study on the input/output resolutions of our model}. 
  \textdagger Timings for the highest resolution models comprise 10 epochs training of the \mr~model and 5 epochs of the \hr~model. }
\label{tab:resolutions}
\end{table*}

\begin{figure*}[!ht]
  \centering
  \resizebox{\textwidth}{!}{
  \newcommand{\turnheightnew}{0.175\columnwidth}

\centering

\begin{tabular}{P{0.25\textwidth}P{0.25\textwidth}P{0.25\textwidth}P{0.25\textwidth}}

\textbf{Input} & \textbf{Monodepth2 MS} $128 \times 416$ &  \textbf{Monodepth2 MS} $192 \times 640$ &  \textbf{Monodepth2 MS} $320 \times 1024$ \\

\multicolumn{4}{c}{\includegraphics[width=1.0\textwidth]{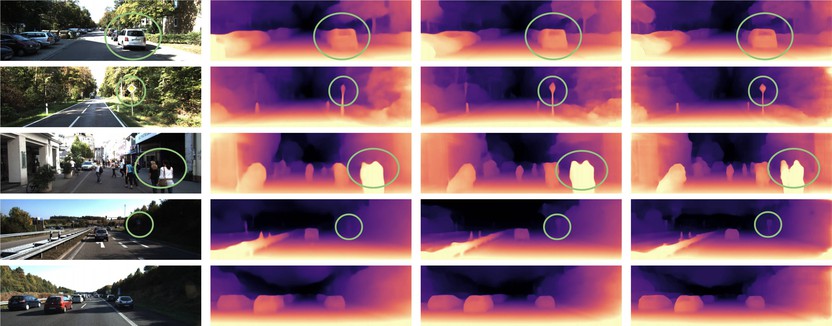}}

\end{tabular}

 }
  \caption{\textbf{Effect of varying resolutions on the KITTI Eigen split.} All predicted disparity maps have been resized to the same size for visualization.
  Our lowest resolution model ($128 \times 416$) captures the broad shape of the scene successfully, but struggles with thin objects and sometimes fails to accurately capture the shape of depth discontinuities around object boundaries.}
  \vspace{-8pt}
  \label{fig:resolutions_qual_resolution}    
\end{figure*}

\section{Results with Post-Processing}
\label{sec:post_processing}
Post-processing, introduced by \cite{godard2017unsupervised}, is a technique to improve test time results on stereo-trained monocular depth estimation methods by running each test image through the network twice, once unflipped and then flipped. 
The two predictions are then masked and averaged. 
This has been shown to bring significant gains in accuracy for stereo results, at the expense of requiring two forward-passes through the network at test time \cite{godard2017unsupervised, poggi20183net}.
In Table~\ref{tab:kitti_eigen_pp} we show, for the first time, that post-processing also improves quantitative performance in the monocular only (M) and mixed (MS) training cases.

\newcommand{\clinegray}{\arrayrulecolor{light-gray}\cline{2-11}\arrayrulecolor{black}}
\newcommand{\clineblack}{\arrayrulecolor{black}\cline{2-11}\arrayrulecolor{black}}

\begin{table*}[h!t]
\centering
\resizebox{0.9\textwidth}{!}{
\begin{tabular}{l|l|c|c||c|c|c|c|c|c|c|}
\clineblack
& Pose network architecture                    & Input frames & Pretrained &

\cellcolor{col1}Abs Rel & \cellcolor{col1}Sq Rel & \cellcolor{col1}RMSE  &
      \cellcolor{col1}\begin{tabular}{@{}c@{}}RMSE \\ log\end{tabular} &
      \cellcolor{col2}$\delta < $1.25 & \cellcolor{col2}$\delta < $1.25$^{2}$ & \cellcolor{col2}$\delta <$ $1.25^{3}$ \\

\clineblack
 & PoseCNN \cite{wang2017learning} & 2 & \checkmark &
0.138 &   1.122 &   5.308 &   0.209 &   0.840 &   0.950 &   0.978 \\  %
 & PoseCNN \cite{wang2017learning} & 3 & \checkmark &
0.148 &   1.211 &   5.595 &   0.219 &   0.815 &   0.942 &   0.976 \\  %
\clinegray
 & Shared encoder (\emph{arXiv v1}) & 2 &  \checkmark &
0.125 &   0.986 &   5.070 &   0.201 &   0.857 &   0.954 &   0.979 \\ %
 & Shared encoder (\emph{arXiv v1})  & 3 &  \checkmark &
0.123 &   1.031 &   5.052 &   0.199 &   0.863 &   0.954 &   0.979 \\ %
\clinegray
 \textbf{Monodepth2} $\Rightarrow$ & Separate ResNet & 2 &   \checkmark&
\textbf{0.115} & 0.919 & 4.863 & \textbf{0.193} & \textbf{0.877} & 0.959 & \textbf{0.981} \\ %
 & Separate ResNet & 3 & \checkmark  &
\textbf{0.115} & \textbf{0.902} & \textbf{4.847} & \textbf{0.193} & \textbf{0.877} & \textbf{0.960} & \textbf{0.981} \\ %
\clineblack

 & PoseCNN \cite{wang2017learning} & 2 &  &
0.147 &   1.164 &   5.445 &   0.221 &   0.818 &   0.940 &   0.974 \\ %
 & PoseCNN \cite{wang2017learning} & 3 &  &
0.147 &   1.117 &   5.403 &   0.222 &   0.815 &   0.940 &   0.976 \\ %
\clinegray
 & Shared encoder (\emph{arXiv v1}) & 2 &   &
0.149 &   1.153 &   5.567 &   0.229 &   0.807 &   0.934 &   0.972 \\ %
 & Shared encoder (\emph{arXiv v1}) & 3 &   &
0.145 &   1.159 &   5.482 &   0.224 &   0.818 &   0.937 &   0.973 \\ %
\clinegray
\textbf{Monodepth2} $\Rightarrow$ & Separate ResNet & 2 &  &
\textbf{0.132} &   1.044 &   \textbf{5.142} &   {\bf 0.210} &    \textbf{0.845} &  \textbf{0.948} &   \textbf{0.977} \\  %
& Separate ResNet & 3 &   &
\textbf{0.132} &   \textbf{1.017} &   5.169 &   0.211 &   0.842 &   0.947 &   \textbf{0.977} \\ %
\clineblack
\end{tabular}}
\vspace{3pt}
\caption{\label{tab:pose_encoder_comparison}\textbf{Ablation of the effect of pose networks on depth prediction.} Results shown are on depth prediction on the KITTI dataset, when trained from monocular sequences only. `Input Frames' indicate how many frames are fed to the pose network. `Shared encoder (\emph{arXiv v1})' denotes the architecture proposed in \emph{v1} of this paper.}
\end{table*}

\section{Effect of Image Resolution}
\label{sec:effect_of_resolution}

In the main paper, we presented results at our standard resolution ($640 \times 192$).
We also showed additional results at higher ($1024 
\times 320$) and lower ($416 \times 128$) resolutions.
In Table \ref{tab:resolutions} we show a full set of results at all three resolutions.
We see that higher resolution helps, confirming the finding in \cite{pillai2018superdepth}.
We also include an ablation showing that, even at the highest resolution, our full-res multi-scale still provides benefit beyond just higher resolution training (\vs `\textbf{Ours} w/o full-res multi-scale').

Our high resolution models were initialized using the weights from our standard resolution ($640 \times 192$) model after 10 epochs of training.
We then trained our high resolution models for 5 epochs with a learning rate of $10^{-5}$.
We used a batch size of 4 to enable this higher resolution model to fit on a single 12GB Titan X GPU.

Qualitative results of the effect of resolution are illustrated in Fig.~\ref{fig:resolutions_qual_resolution}. 
It is clear that all resolutions accurately capture the overall shape of the scene.
However, only the highest resolution model accurately represents the shape of thin objects.

\section{Comparison of Pose Encoder}
\label{sec:pose_encoder_comparison}
In Table~\ref{tab:pose_encoder_comparison} we evaluate different pose encoders.
In an earlier version of this paper, we proposed the use of a shared pose encoder that shared features with the depth network. 
This resulted in fewer parameters to optimize during training, but also results in a decrease in depth prediction accuracy, see Table~\ref{tab:pose_encoder_comparison}.
As a baseline we compare against the pose network used by \cite{wang2017learning}, which builds upon \cite{zhou2017unsupervised} with an additional scaling of the translation by the mean of the inverse depth.
Overall, our separate encoder is superior for both pretrained and non-pretrained variants, whether we use two or three frames as input.

\section{Supplementary Video Results}
\begin{figure}
  \centering
  \includegraphics[width=\columnwidth]{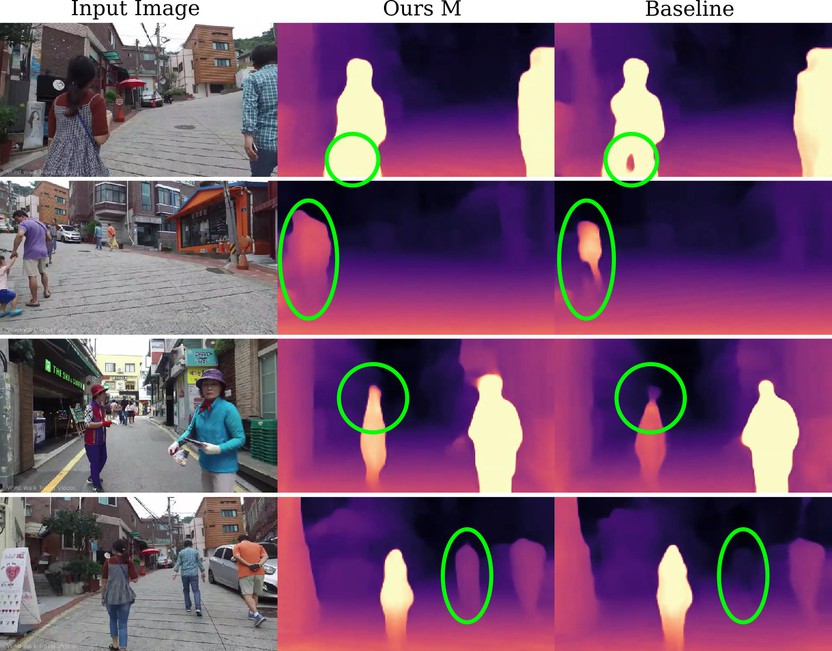}
  \caption{\textbf{Additional Wander results.} We observe that our model (Ours M) results in fewer visual artifacts when compared to the the baseline (\ie the same model including VGG loss, but without our contributions).}
  \label{fig:wander_qual}    
\end{figure}

In the supplementary video, we show results on `Wander', a monocular dataset collected from the `Wind Walk Travel Videos' YouTube channel.\footnote{\url{https://www.youtube.com/channel/UCPur06mx78RtwgHJzxpu2ew}}
This dataset is quite different from the car mounted videos of KITTI as it only features a \emph{monocular} hand-held camera  in a non-European environment. 
We train on four sequences and present results on a fifth unseen sequence.
We use an input/output resolution of $128 \times 224$.
As with our KITTI experiments we train for 20 epochs with a batch size of 12, with a learning rate of $10^{-4}$ which is reduced by a factor of 10 for the final 5 epochs.
For these handheld videos we found that the SSIM loss produced artifacts at object edges.
As a result, we used a feature reconstruction loss in the appearance matching term, as in \cite{ren2017unsupervised,sun2017pwc,zhanst2018}, by computing the L1 distance on the reprojected and normalized \texttt{relu1\_1} features from an \mbox{ImageNet} pretrained VGG16~\cite{simonyan2014very} as our $pe$ function. This takes significantly longer to train, but results in qualitatively better depth maps on this dataset.
Examples of predicted depths can be seen in Fig.~\ref{fig:wander_qual}.

\end{appendices}

\end{document}